%% file: main.tex
\documentclass{article}
\usepackage{amssymb}

\usepackage[final]{corl_2025} 

\usepackage{hyperref}

\usepackage{amsmath}
\usepackage{amssymb}
\usepackage{mathtools}
\usepackage{amsthm}
\usepackage{booktabs}

\usepackage{wrapfig}

\usepackage[capitalize,noabbrev]{cleveref}

\theoremstyle{plain}

\theoremstyle{definition}

\theoremstyle{remark}

\usepackage[textsize=tiny]{todonotes}

\usepackage{array} 
\usepackage{tabularx} 
\usepackage{float}
\usepackage{enumitem}

\usepackage[format=plain,
            labelfont=it,
            labelsep=period]{caption}

\definecolor{niceblue}{RGB}{0,149,255}
\definecolor{niceorange}{RGB}{180,95,6}
\newcommand{\thewebsite}{\href{https://aalmuzairee.github.io/mad}{\color{niceblue}\bf aalmuzairee.github.io/mad}}

\title{Merging and Disentangling Views in Visual Reinforcement Learning for Robotic Manipulation}

\author{
  Abdulaziz Almuzairee \quad Rohan Patil \quad Dwait Bhatt \quad Henrik I. Christensen \vspace{0.05in}\\ 
  UC San Diego \vspace{0.14in}\\ 
  \large{\thewebsite} 
}

\begin{document}
\maketitle

\begin{figure}[H]
    \centering
    \includegraphics[width=0.9\linewidth,trim={0.cm 0.2cm 0.0cm 0.1cm},clip]{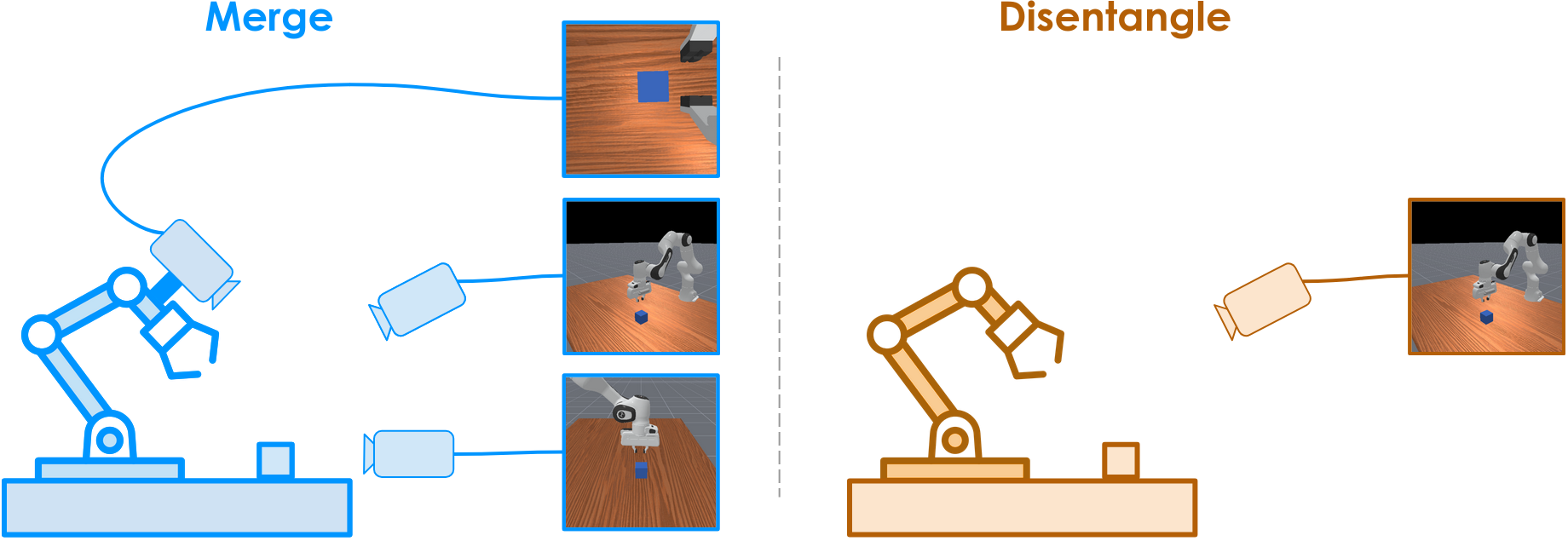}
    
    \vspace{0.3cm}
   \caption{\textbf{Merge And Disentangle.} We introduce a method that can merge multiple camera views during training to learn better representations, while simultaneously disentangling the camera view representations, such that the policy can function with any singular view input during deployment.}
   \label{fig:head_diag}
\end{figure}


\begin{abstract}
Vision is well-known for its use in manipulation, especially using visual servoing. Due to the 3D nature of the world, using multiple camera views and merging them creates better representations for Q-learning and in turn, trains more sample efficient policies. Nevertheless, these multi-view policies are sensitive to failing cameras and can be burdensome to deploy. To mitigate these issues, we introduce a \textbf{M}erge \textbf{A}nd \textbf{D}isentanglement (\textbf{MAD}) algorithm that efficiently merges views to increase sample efficiency  while simultaneously disentangling views by augmenting multi-view feature inputs with single-view features. This produces robust policies and allows lightweight deployment. We demonstrate the efficiency and robustness of our approach using Meta-World and ManiSkill3.  

\end{abstract}



\section{Introduction}

Visual reinforcement learning (RL) has demonstrated remarkable progress, achieving superhuman performance on Atari games~\cite{mnih2013playing} and successfully tackling complex real-world applications from robotic manipulation to autonomous flight~\cite{levine2016end, pinto2016supersizing, kalashnikov2018qt, kaufmann2023drone}. However, the fundamental limitation of learning from 2D observations constrains these systems' understanding of the 3D world, particularly in robotics where depth perception and control are differentiators.

Multi-view approaches offer promising solutions for visual RL, especially when combined with data efficient Q-learning algorithms. They can \emph{merge} multiple views to learn better representations, overcome occlusions, and achieve higher sample efficiency~\cite{hsu2022vision}. However, they come with a practical challenge: conditioning policies on multi-view representations can be burdensome to deploy and render them fragile in the case of malfunctioning sensors. To deploy lightweight policies that are robust to a reduction in available camera views, policies must be carefully \emph{disentangled} during training \cite{dunion2024multi}.

Aligning these two directions, we propose a method to \textbf{M}erge \textbf{A}nd \textbf{D}isentangle views (\textbf{MAD}), thereby leveraging multi-view representations for \emph{improved sample efficiency}, while being \emph{robust to a reduction of camera views} for lightweight deployment. To accomplish that, MAD processes each camera view input individually through a single shared CNN encoder. Then, all singular view features are merged through feature summation to create a merged multi-view feature representation. This merged feature representation is passed down to the downstream actor (\emph{policy}) and critic (\emph{$Q$-function}) for learning. However, to properly disentangle view features, the downstream actor and critic need to also train on the singular view features. Naively training on both merged feature representations and all their singular view representations decreases sample efficiency and destabilizes learning, as they can be viewed as multiple different states by the downstream actor and critic networks. Therefore, we build upon SADA \cite{almuzairee2024recipe}, a framework for applying data augmentation to visual RL agents that stabilizes both the actor and the critic under data augmentation by \emph{selectively} augmenting their inputs. We modify the RL loss objectives, such that we train on each merged multi-view feature, and apply all its singular view features as augmentations to it. By using this formulation, we are able to increase sample efficiency while ensuring robustness to a reduction in input camera views. Most importantly, we achieve this without requiring ordered input views, auxiliary losses, extra forward passes, or additional learnable parameters.

Methods that merge and disentangle are vital when deploying robots to new environments or designing benchmarks, where determining optimal camera placement and ensuring task observability is non-trivial. By learning to merge information from multiple views while maintaining the ability to generalize to individual perspectives, policies can leverage the complementary nature of different viewpoints, benefiting from the increased sample efficiency and ensuring robustness. This process of merging multiple camera views and simultaneously disentangling individual views to create robust policies is illustrated in Figure \ref{fig:head_diag}.

We evaluate our method on 20 visual RL tasks from Meta-World \cite{yu2020meta} and ManiSkill3 \cite{tao2024maniskill3} benchmarks, and find that MAD is able to achieve higher sample efficiency than baselines while being robust to a reduction in available camera views.

\begin{figure}[t]
    \centering
    \includegraphics[width=\linewidth,trim={0 0.5cm 0 0.3cm},clip]{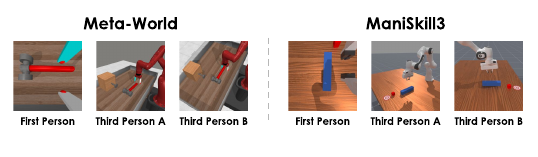}
   \caption{\textbf{Environment Setup.} Our environment setup for robotic manipulation where we use three input camera views as inputs, namely First Person, Third Person A, and Third Person B across two visual RL benchmarks: \emph{(Left)} Meta-World \emph{(Right)} ManiSkill3. Extended visuals in Appendix \ref{sec:app_visuals}.}
   \label{fig:main_cam_setup}
\end{figure}

\section{Related Works}

\textbf{Data Augmentation in Visual Reinforcement Learning.}
To improve the visual generalization of models, data augmentation has been a commonly employed strategy in supervised and self-supervised learning for computer vision tasks~\cite{noroozi2016unsupervised, wu2018unsupervised, oord2018representation, tian2019contrastive, chen2020simple, he2022masked}. Due to its limited data diversity, visual reinforcement learning is especially prone to overfitting on visual inputs. Recent works have found that applying data augmentation with input image transformations such as random crops or random shifts regularizes the learning and increases sample efficiency in RL agents~\cite{srinivas2020curl, laskin2020reinforcement, kostrikov2020image, stooke2020atc, yarats2021reinforcement, yarats2021mastering}. In contrast, \citep{laskin2020reinforcement, Raileanu2020AutomaticDA} find that stronger image transformations such as rotation, random convolution, and masking lead to training instabilities and a decrease in data efficiency. To mitigate this, many works focus on improving training stability under stronger data transformations ~\cite{Raileanu2020AutomaticDA, hansen2021stabilizing, Wang2021UnsupervisedVA, fan2021secant, yuan2022don, grooten2023madi, yang2024movie, almuzairee2024recipe} and increasing visual generalization by proposing new types of transformations ~\cite{lee2019sample, wang2020improving, hansen2021softda, zhang2021generalization, huang2022spectrum, wang2023generalizable, teoh2024green}. Most importantly, \citet{almuzairee2024recipe} propose a generic recipe for applying data augmentation, named SADA, which allows RL agents to generalize to many types of stronger image transformations, without sacrificing training sample efficiency.  Our work builds on SADA and shows that it can be extended to feature level data augmentation within our framework.

\textbf{Robot Manipulation from Multiple Views.} 
While robots can learn from singular camera views, increasing the available cameras and using merged representations have been shown to improve performance in robotic manipulation. ~\cite{akinola2020learning, hsu2022vision, seo2023multi, gervet2023act3d, goyal2024rvt2, qian20243d, yuan2024learning}. These approaches leverage multiple camera perspectives to mitigate occlusions and improve representations, leading to enhanced task performance \cite{szot2021habitat, hsu2022vision}. Most notably, \citet{hsu2022vision} show that a first-person view outperforms a third-person view when considering only one single view. However, due to the first-person view's limited observability, fusing it with a third person camera improves learning, as long as the third person view is regularized with a variational information bottleneck. They show that their method, VIB, outperforms singular and combined views on tabletop robot manipulation. \citet{jangir2022look} achieve similar conclusions, where fusing first and third person views with cross attention achieves better real-world transfer than their singular view counterparts. Although there is merit in using multiple views, performance degradation is significant when a view that was available during training is not present during inference, especially if the views were not disentangled properly during training~\cite{dunion2024multi}.

\textbf{Merging and Disentangling Features.} 
Research in computer vision has extensively explored feature merging ~\cite{su2015multi, yao2018mvsnet, borse2023xalign, li2024bevformer} and feature disentanglement ~\cite{oord2018representation, khosla2020supervised, oquab2023dinov2} across multiple camera views and sensor modalities. Inspired by these efforts, multiple works in visual RL have explored feature merging ~\cite{li2019multi, yang2022self, jangir2022look, hsu2022vision} and feature disentanglement ~\cite{eysenbach2018diversity, srinivas2020curl, park2023metra}. However, policies trained on multiple feature inputs often suffer significant performance degradation when any feature inputs become unavailable ~\cite{vasco2021sense, dunion2024multi}. Multiple works have researched this problem of feature disentanglement for reinforcement learning. \citet{hu2024privileged} explore the use of scaffolding through extra privileged features at training time, and disentangling them to achieve better test time performance than their unscaffolded counterparts. \citet{skand2024simple} introduce a multi sensor feature encoder for RL polices that disentangles features through dropout, yielding robust policies. Recently, \citet{dunion2024multi} addressed feature disentanglement in the multi-camera view RL setup for robotic manipulation. They introduced MVD, an actor-critic method that \emph{disentangles} view representations into shared and private components through contrastive losses, enabling policies to maintain performance despite a reduction in input camera views. While MVD focuses exclusively on disentangling views, our approach focuses on both merging and disentangling views: we \emph{merge} multi-view features to improve sample efficiency and we train on both merged multi-view features and singular-view features to ensure \emph{disentanglement} in the case of a singular camera view input.

\section{Preliminaries}

\textbf{Visual Reinforcement Learning} formulates the interaction between the agent and the environment as a Partially Observable Markov Decision Process (POMDP) \cite{kaelbling1998pomdp}, defined by a tuple of $\langle \mathcal{S}, \mathcal{O}, \mathcal{A}, \mathcal{T}, R, \gamma \rangle$. In a POMDP,  $\mathcal{S} \subseteq \mathbb{R}^d$ is an unobservable state space by the agent, ${o} \in \mathcal{O}$ is the observation space, ${a} \in \mathcal{A}$ the action space, $\mathcal{T}: \mathcal{S} \times \mathcal{A} \times \mathcal{S} \rightarrow [0, 1]$ is the state-to-state transition probability function, $R: \mathcal{S} \times \mathcal{A} \rightarrow \mathbb{R}$ is the reward function and $\gamma$ is the discount factor. To better approximate the current state $s_t \in \mathcal{S}$, we follow \citet{mnih2013playing} in defining observations as a stack of three prior consecutive RGB frames at time $t$. The goal is to learn a policy $\pi: \mathcal{O} \rightarrow \mathcal{A}$ that maximizes the expected discounted sum of rewards $\mathbf{E}_\pi [\sum_{t=0}^\infty \gamma^t r_t]$ where $r_t =  R({s}_t, {a}_t)$.

\textbf{Data Regularized Q-Learning} (DrQ) \cite{kostrikov2020image} is an actor critic algorithm, based on Soft Actor Critic \cite{haarnoja2018soft} commonly used for continuous control in visual RL. DrQ concurrently learns a Q-function $Q_\theta$ (critic) and a policy $\pi_\phi$ (actor) with a seperate neural network for each. The Q-function $Q_{\theta}$ aims to estimate the optimal state-action value function $Q^{*} \colon \mathcal{O} \times \mathcal{A} \mapsto \mathbb{R}$ by minimizing the one step Bellman Residual  $\mathcal{L}_{Q_{\theta}}(\mathcal{D}) = \mathbb{E}_{(\mathbf{o}_{t},\mathbf{a}_{t}, r_{t}, \mathbf{o}_{t+1})\sim\mathcal{D}} [ ( Q_\theta(\mathbf{o}_{t}, \mathbf{a}_{t}) -
r_t - \gamma Q_{\overline{\theta}}(\mathbf{o}_{t+1}, \mathbf{a}') )^2 ]$ where $\mathcal{D}$ is the replay buffer, $\mathbf{a}'$ is an action sampled from the current policy $\mathbf{a}' \sim \pi_\phi(\cdot |\mathbf{o}_{t+1})$, and $Q_{\overline{\theta}}$ represents an exponential moving average of the weights from $Q_\theta$ \citep{watkins1992q,lillicrap2015continuous,haarnoja2018soft}. The policy $\pi_\phi$ is a stochastic policy with temperature alpha $\alpha$ that aims to maximize entropy and $Q$-values. For simplicity, we abstract the entropy objective and define a generic actor loss to be $ \mathcal{L}_{\pi_\phi}(\mathcal{D}) = \mathbb{E}_{\mathbf{o}_{t} \sim \mathcal{D}}  \left[-Q_\theta(\mathbf{o}_t, \pi_\phi(\mathbf{o}_t)) \right]$. Both the critic and actor losses are updated iteratively using stochastic gradient descent in an aim to maximize the expected discounted sum of rewards. DrQ also applies small random shifts to all input images sampled from the replay buffer which improves learning sample efficiency \cite{kostrikov2020image, yarats2021mastering}.

\begin{figure*}[t]
    \centering
    \includegraphics[width=\linewidth,trim={0 0cm 0 0cm},clip]{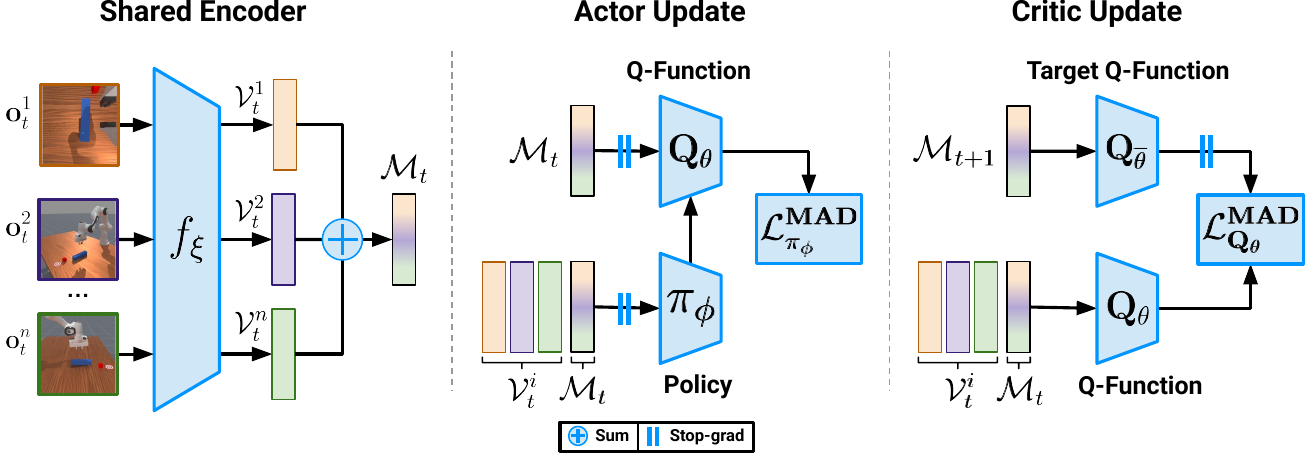}
    \caption{\textbf{Our framework.} Update diagram of a generic visual actor-critic model with our modifications. MAD \emph{merges} camera views through feature summation and \emph{disentangles} camera views by selectively augmenting inputs to the downstream actor and critic with all the singular view features. The agent is trained end-to-end with our defined MAD loss functions. \emph{(Left)}: Single Shared CNN Encoder. \emph{(Middle)}: Actor Update diagram  \emph{(Right)}: Critic Update diagram. }
    \label{fig:main_arch}
\end{figure*}

\section{Merge And Disentangle Views in Visual Reinforcement Learning}
\vspace{-0.04in}

Building on prior work in multi-view reinforcement learning, we found it to pursue two distinct directions: one focusing on merging views to improve sample efficiency, while the other focusing on disentangling views to create policies that are robust to view reduction. To unify these complementary approaches in pursuit of both increased sample efficiency and policy robustness, we propose \textbf{MAD}: \textbf{M}erge \textbf{A}nd \textbf{D}isentangle views for visual reinforcement learning. Our method merges views through feature summation, and simultaneously disentangles views by training the downstream actor and critic on both the merged and singular view features. However, \emph{naively} training on both merged and singular view features deteriorates policy performance. Therefore, we elect to train using the merged features, and apply singular view features as augmentations to the inputs of the downstream actor and critic networks. We start by defining our merge module, followed by our application of feature-level data augmentation, and finally our augmentation strategy.

\vspace{-0.04in}
\subsection{Merge Module}
\vspace{-0.04in}

Given a multi-view input $\mathbf{o}^{m}_t = {\mathbf{o}^1_t, \mathbf{o}^2_t, ..., \mathbf{o}^n_t}$ consisting of $n$ views at time $t$, where each $\mathbf{o}^i_t$ represents a single view, each view is passed separately through a shared CNN encoder $f_\xi$. The output features of a single view input is then defined to be $\mathcal{V}^i_t=f_\xi(\mathbf{o}^i_t)$. After encoding all singular views $ {\mathcal{V}^1_t, \mathcal{V}^2_t, ..., \mathcal{V}^n_t}$, the features are merged through summation such that the combined multi-view representation becomes: $\mathcal{M}_t = \sum_i^n{\mathcal{V}^i_t} $.

Feature summation is chosen as the merging method for two reasons: 1) To make the downstream actor and critic robust to a reduction in views, they need to be trained on both merged features $\mathcal{M}_t$ and singular-view features $\mathcal{V}^i_t$. With feature summation, both $\mathcal{M}_t$ and $\mathcal{V}^i_t$ have equal dimensionality, and thus no architectural changes need to be made to accommodate different feature dimensions in the case of missing input views. 2)  Feature summation preserves the magnitude of different view features, such that the downstream actor and critic have a signal of how many views are inputted. As our experiments will show, most merging methods in visual RL perform similarly, and so the choice of the merging method comes down to the properties desired within a certain framework.

\vspace{-0.04in}
\subsection{Feature-level Data Augmentation}
\vspace{-0.04in}

Feature-level augmentation in MAD differs from traditional RL data augmentation techniques. While conventional approaches apply data augmentation by modifying input images through random cropping and color jittering, or altering input states via amplitude scaling and Gaussian noise injection \cite{laskin2020reinforcement}, MAD introduces augmentation at the feature level—specifically between the image encoder and the downstream actor and critic networks. The process begins by encoding each camera view image into its corresponding singular-view feature representation $\mathcal{V}^i_t$, then combining these features through summation to produce the merged representation $\mathcal{M}_t$. Standard multi-view algorithms would only pass this merged representation $\mathcal{M}_t$ to the downstream components. However, to strengthen the robustness of the downstream actor and critic to missing input camera views, MAD augments the downstream component inputs with all the singular-view feature representations $\mathcal{V}^i_t$.

\vspace{-0.04in}
\subsection{Augmentation Strategy}

To generalize to all singular view inputs $\mathbf{o}^i_t$, MAD applies their corresponding features ${V}^i_t$ as feature-level augmentations to the downstream actor and critic. However, naive application of data augmentation in visual RL has been shown to degrade policy performance and stability \cite{laskin2020reinforcement, Raileanu2020AutomaticDA, almuzairee2024recipe}. Later works established ways to stabilize actor-critic learning under data augmentation \cite{hansen2021stabilizing,almuzairee2024recipe}. We follow SADA \cite{almuzairee2024recipe} in their recipe for applying data augmentation, where given an encoder $f_\xi$, actor $\pi_\phi$, and critic $Q_\theta$, data augmentation is \emph{selectively} applied to the actor and critic inputs such that: \emph{(1)} In the critic update, the target $Q$-value is predicted \emph{purely} from the unaugmented stream, while the online $Q$-value is predicted from \emph{both} the augmented and unaugmented streams. \emph{(2)} In the actor update, the $Q$-value is predicted \emph{purely} from unaugmented stream while the policy action is predicted from \emph{both} the augmented and unaugmented streams. By predicting the learning targets in both actor and critic updates from the unaugmented stream, the variance in the targets is reduced, thereby stabilizing the learning objective under data augmentation.

We build on this loss formulation with some modifications. Given $n$ input camera views, $n$ single-view features $\mathcal{V}^{i}_{t}$ = ${f_\xi(\mathcal{\mathbf{o}}^{i}_{t})}$ for $i \in \{1,...,n\}$, and a multi-view feature representation $\mathcal{M}_t$ = $\sum_{i=1}^n{\mathcal{V}^{i}_t}$, the MAD update loss function for generic actor becomes:

\vspace{-0.24in}

\begin{align}
   \mathcal{L}_{\pi_\phi}^\textbf{UnAug}(\mathcal{D}) &= \mathbb{E}_{\mathbf{o}^{m}_{t} \sim \mathcal{D}} \notag  \left[-Q_\theta(\mathcal{M}_t, \pi_\phi(\mathcal{M}_t)) \right] \\
   \mathcal{L}_{\pi_\phi}^\textbf{Aug}(\mathcal{D}, i) &= \mathbb{E}_{\mathbf{o}^{m}_{t} \sim \mathcal{D}} \notag  \left[-Q_\theta(\mathcal{M}_t, \pi_\phi(\mathcal{V}^{i}_t)) \right] \\
   \mathcal{L}_{\pi_\phi}^\textbf{MAD}(\mathcal{D}) = \alpha *  \mathcal{L}_{\pi_\phi}^\textbf{UnAug}&(\mathcal{D}) \; + \; (1-\alpha) * \frac{1}{n} \sum_{i=1}^n\mathcal{L}_{\pi_\phi}^\textbf{Aug}(\mathcal{D}, i) ~~~\textcolor{gray}{\textrm{(actor)}}
\end{align}
\vspace{-0.1in}

where $\alpha$ is a hyperparameter that we add to the SADA loss. This $\alpha$ hyperparameter weighs the unaugmented and augmented streams for more fine grained control over the learning, similar to \cite{hansen2021stabilizing}. Setting this $\alpha$ hyperparameter to 0.5 recovers the original SADA loss. On the other hand, the MAD update loss function for a generic critic becomes:
\vspace{-0.04in}
\begin{align}
   \mathcal{L}^\textbf{UnAug}_{Q_{\theta}}(\mathcal{D}) = \mathbb{E}_{(\mathbf{o}^m_{t},\mathbf{a}_{t}, r_{t}, \mathbf{o}^m_{t+1})\sim\mathcal{D}}& \Big[ \big( Q_\theta(\mathcal{M}_t, \mathbf{a}_{t}) \;-
    r_t - \gamma Q_{\overline{\theta}}(\mathcal{M}_{t+1}, \mathbf{a'}) \big)^2 \Big] \notag \\
   \mathcal{L}^\textbf{Aug}_{Q_{\theta}}(\mathcal{D}, i) = \mathbb{E}_{(\mathbf{o}^m_{t},\mathbf{a}_{t}, r_{t}, \mathbf{o}^m_{t+1})\sim\mathcal{D}}& \Big[ \big( Q_\theta(\mathcal{V}^{i}_t, \mathbf{a}_{t}) \;- r_t - \gamma Q_{\overline{\theta}}(\mathcal{M}_{t+1}, \mathbf{a'}) \big)^2 \Big] \notag \\
   \mathcal{L}_{Q_{\theta}}^\textbf{MAD}(\mathcal{D}) = \alpha *  \mathcal{L}_{Q_{\theta}}^\textbf{UnAug}&(\mathcal{D}) \; + \; (1-\alpha) * \frac{1}{n} \sum_{i=1}^n\mathcal{L}_{Q_\theta}^\textbf{Aug}(\mathcal{D}, i) ~~~\textcolor{gray}{\textrm{(critic)}}
\end{align}
A higher value of $\alpha$ would increase the weight of the unaugmented objective, while a lower $\alpha$ would increase the weight of the augmented objective. Using this formulation, and after tuning $\alpha$, MAD is able to train on both the merged features and singular view features with minimal loss to training sample efficiency.  A detailed diagram of our method update is provided in Figure \ref{fig:main_arch}.

\begin{figure*}[t]
    \centering
    \includegraphics[width=\textwidth,trim={0 0 0 2.3cm},clip]{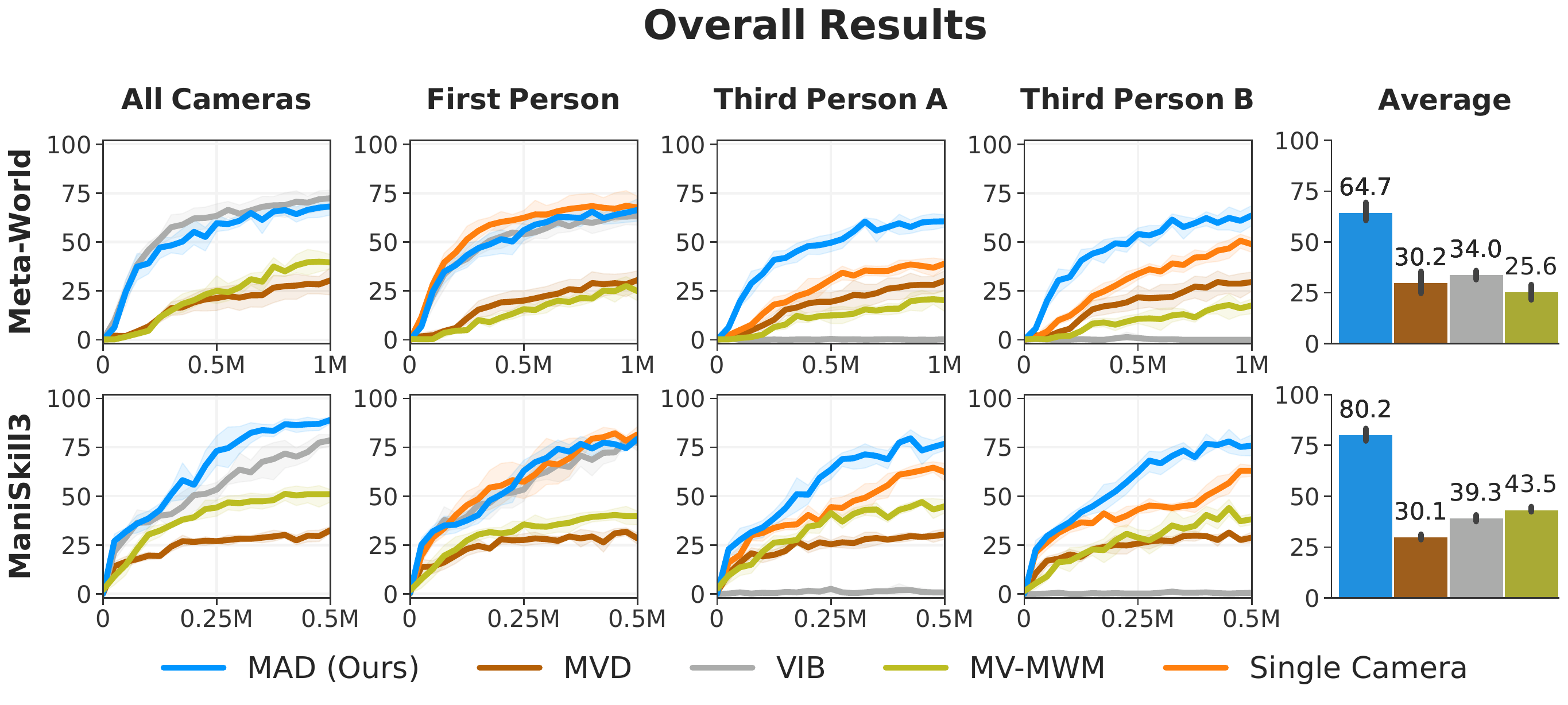}
    \vspace{-0.2in}
    \caption{\textbf{Overall Robustness.} Success rate as a function of environment steps, averaged over all \emph{(Top)} 15 Meta-World and \emph{(Bottom)} 5 ManiSkill3 visual RL tasks. Methods are trained on all three camera views and evaluated on all and singular camera views with the final average displayed on the far right. Mean and 95\% CI over 5 random seeds. }
    \label{fig:main_exp}
    \vspace{-0.1in}
\end{figure*}

\section{Experiments}

We benchmark our method and baselines on a total of 20 visual RL tasks, consisting of 15 Meta-World v2 tasks~\cite{yu2020meta} and 5 ManiSkill3 tasks~\cite{tao2024maniskill3}, focusing on tabletop robot manipulation tasks. Both environments are set up with three camera views, consisting of a first-person camera attached to the robot arm, and two third-person cameras with reasonable observability to the task at hand. See Figure \ref{fig:main_cam_setup} for visualizations of our camera setup in each environment. The full task list is defined in \ref{subsection:env_setups}, and detailed visuals of all tasks can be found in \ref{sec:app_visuals}. Through experimentation, our aim is to answer the following questions:

\vspace{-0.08in}
\begin{itemize}[label={-},leftmargin=0.25in]
    \setlength\itemsep{-0.1em}
    \item \textbf{Robustness.} How does MAD compare to baselines in terms of sample efficiency? Does MAD function with a reduction in camera views? 
    \item \textbf{Analysis.} Why do baselines fail to achieve similar sample efficiency to MAD? How much does each component contribute within MAD? How do different merge methods compare? 
    \item \textbf{Adaptability.}  How well does MAD adapt with occluded views? Can MAD be adapted to other modalities? 
\end{itemize}
\vspace{-0.08in}

\textbf{Implementation.} We use DrQ \cite{kostrikov2020image}, a visual based Soft Actor Critic algorithm \cite{haarnoja2018soft}, as the backbone algorithm for our method across both environments, with detailed hyper parameters defined in Appendix \ref{subsection:app_hparams}. To process all input camera views, we use a single shared DQN CNN encoder \cite{mnih2015human}. For our observations, we pass in a stack of three most recent RGB images such that input observations for each view are of shape ($3\times\mathbb{R}^{\mathbf{(3\times84\times84)}})$.  We further pass in proprioceptive robot state input to the actor and critic, similar to prior works \cite{hsu2022vision,dunion2024multi}. When learning from images, DrQ regularizes its learning by applying small random shift transformations to input images, which has been shown to improve performance. We apply identical random shifts to all input camera views at each timestep. 

\textbf{Environments.} We evaluate on 15 Meta-World \cite{yu2020meta} tasks spanning medium, hard, and very hard difficulties, as defined in \cite{seo2023masked}, and 5 tasks from ManiSkill3 \cite{tao2024maniskill3}. For each of the two benchmarks, we use a fixed camera setup consisting of a first person camera view and two third person camera views. When evaluating, we report the mean success rate of the agent across 20 episodes.

\textbf{Baselines.} We compare MAD against the following strong baselines. 1) \textbf{MVD} \cite{dunion2024multi}, a contrastive learning method that learns to \emph{disentangle} multi-view image inputs by encoding a shared and private representation for each view. The shared representation is pushed to align with other views while the private representation is pushed to differ. Through combined contrastive losses, they are able to learn representations that are robust to a reduction in views.  2) \textbf{VIB} \cite{hsu2022vision}, a variational information bottleneck approach that \emph{merges} first person and third person views with a variational information bottleneck objective applied only on third person views to provide complementary information to the first person view. 3) \textbf{MV-MWM} \cite{seo2023multi}, a multi-view masked autoencoder that learns representations across multiple views while using a world model for planning. While MV-MWM uses expert demonstrations to bootstrap its learning, we train it without any expert demonstrations for a fair comparison with our method and baselines. 4) \textbf{Single Camera} DrQ \cite{kostrikov2020image}, where DrQ agents are trained only on a single view and evaluated on that same single view. 

\vspace{-0.08in}
\subsection{Empirical Results}

\textbf{Robustness.} To quantify the \emph{sample efficiency} of MAD, we train all methods on 20 visual RL tasks from Meta-World and ManiSkill3, using all three cameras as input, and evaluate the methods on all cameras views combined, and individual, reporting the combined, individual and average success rates in Figure \ref{fig:main_exp}. As our results indicate, MAD demonstrates superior performance to baselines on Meta-World by (30\%) success rate over 15 tasks, and on ManiSkill3 by (36\%) over 5 tasks. These improvements in success rates within the same number of environment steps as baseline methods directly reflect the enhanced sample efficiency of MAD. 

With a \emph{reduction in camera views}, MAD achieves stronger robustness on the Third Person A and Third Person B singular views than all the baselines, including the Single Camera baseline that was trained solely on these third person views. This indicates that MAD was able to leverage better representations from All Cameras to achieve higher success rates on these third person views. On the First Person View, MAD outperforms most baselines and achieves similar success rates as the Single Camera and VIB baselines. Overall, MAD is able to achieve consistent success rates whether All Cameras are present, or when only singular camera view inputs are available, indicating a robustness to a \emph{reduction in camera views}. Detailed graphs of all results can be found in Appendix \ref{sec:app_res}. Extended experiments on increasing views (5 views), and varying input view resolutions are provided in Appendix \ref{sec:app_scalability}.

\begin{figure*}[t]
    \centering
    \includegraphics[width=\textwidth,trim={0 0 0 2.8cm},clip]{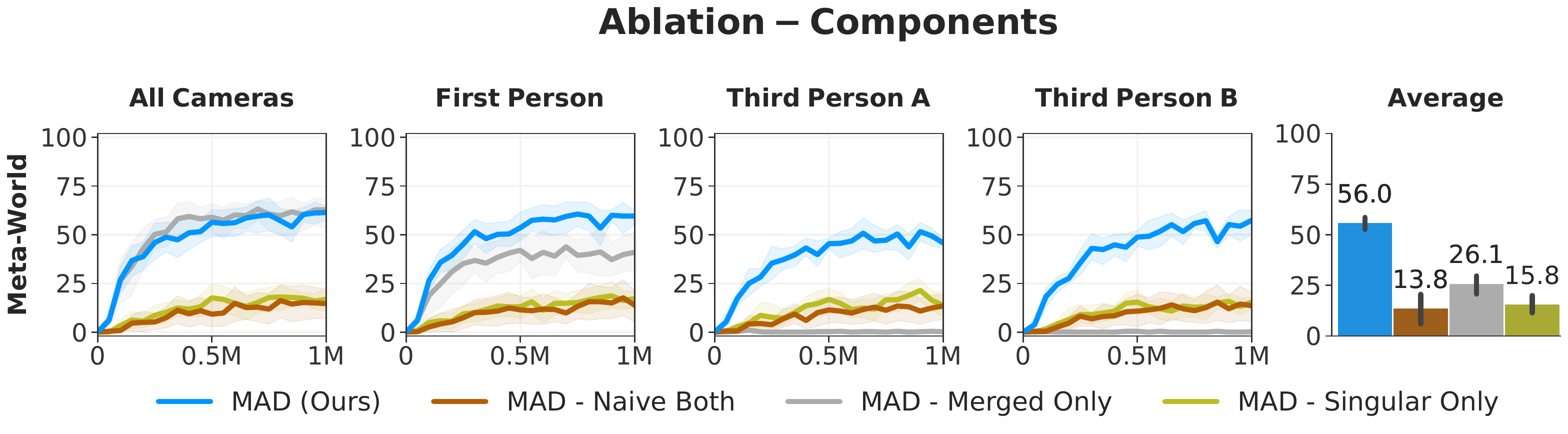}
    \includegraphics[width=\textwidth,trim={0 0 0 2.3cm},clip]{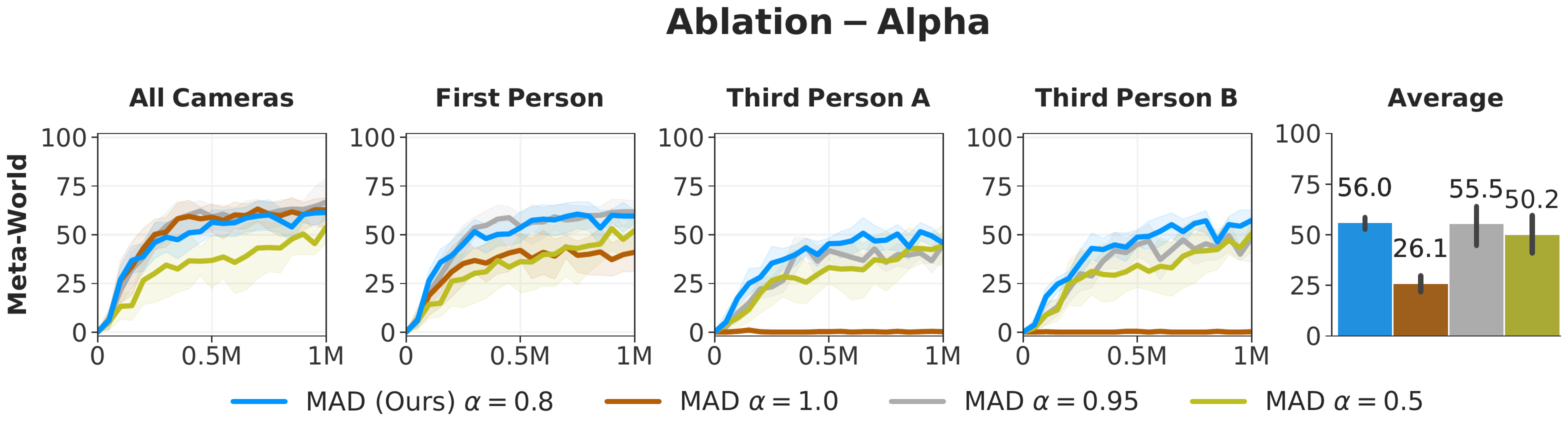}
    \caption{\textbf{Ablations.} Success rate as a function of environment steps, averaged over 5 Meta-World hard tasks. \emph{(Top)} Component Ablations. \emph{(Bottom)} Alpha Ablations. Methods trained on all camera views and evaluated on all and singular camera views. Mean and 95\% CI over 5 random seeds. }
    \label{fig:main_abl}
    \vspace{-0.2in}
\end{figure*} 

\textbf{Analysis.} Based on our experiments, baselines \emph{fail to achieve similar sample efficiency and robustness to MAD} over the two benchmarks.  Our \emph{disentanglement} baseline, MVD, successfully disentangles representations such that it is robust to a reduction in views, but it struggles with poor sample efficiency. This limitation stems from its design approach: rather than leveraging all input views simultaneously, MVD randomly selects individual views to encode into shared and private representations. While this strategy effectively promotes disentanglement, it deliberately avoids creating representations dependent on all views combined — a trade-off that preserves robustness at the cost of efficiency. On the other hand, our \emph{merge} baseline, VIB, appears highly dependent on the first person view, since it applies a variational information bottleneck only on third person views. Therefore, it achieves high sample efficiency only when the first person view is available, and it fails otherwise. Our third baseline, MV-MWM, is able to disentangle views effectively. However, it uses a heavy auxiliary objective of learning masked multi view representations that hinders its learning speed when no expert demonstrations are provided. In contrast to MVD, MAD is able to leverage all input views combined to increase its learning speed, while augmenting with singular view features to disentangle representations. In contrast to VIB, MAD is agnostic to input perspectives, allowing it to function even without first person view inputs. In contrast to MV-MWM, MAD uses no heavy auxiliary objective that requires expert demonstrations to increase its learning speed. Overall, MAD is able to achieve higher sample efficiency and robustness than MVD, VIB, and MV-MWM.

To measure the \emph{contribution of each component within MAD}, we conduct ablations over 5 hard tasks from Meta-World and plot the results in Figure \ref{fig:main_abl}. We first ablate different components. In (\emph{MAD - Naive Both}), we train our DrQ baseline naively on both merged and singular view features without using the \emph{MAD loss formulations}. In (\emph{MAD - Merged Only}), we train our DrQ baseline on the merged multi view features only without any \emph{disentanglement}, and in (\emph{MAD - Singular Only})  we train our DrQ baseline on all the singular view features without \emph{merging} them. MAD outperforms all these alternative setups by (29\%) success rate, indicating the necessity of each component within our formulation. We further ablate different $\alpha$ values for the loss, where we find $\alpha = 0.8$ to outperform other alpha values, including the original SADA formulation of $\alpha = 0.5$. In summary, each of our design choices appears to be essential for the superior performance of MAD.

\begin{figure}[t]
    \centering
    \begin{minipage}[t]{0.48\textwidth}
        \centering
        \includegraphics[width=1.0\textwidth,trim={0cm 1.1cm 0 0.2cm},clip]{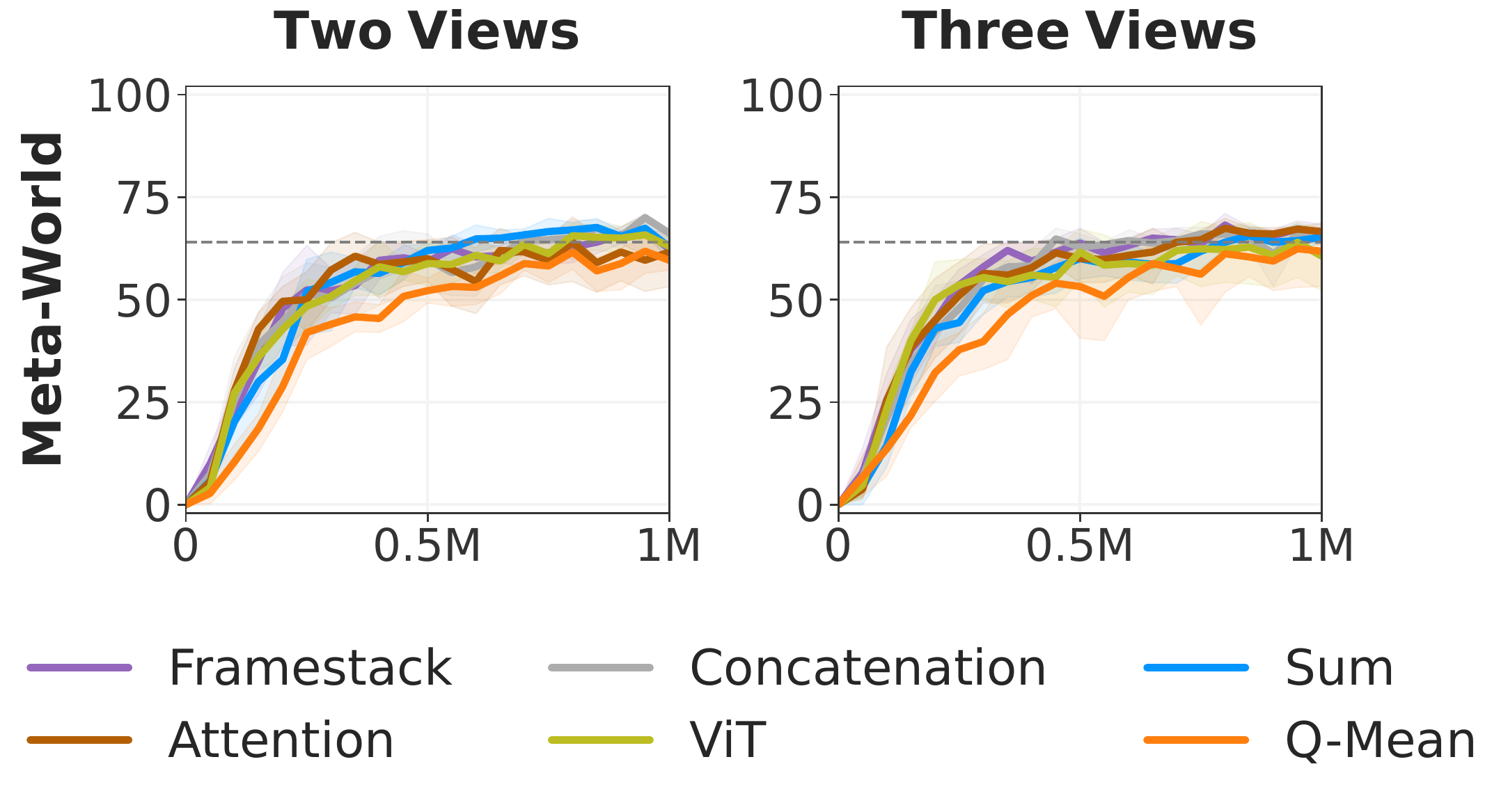}
        \caption{\textbf{Multi-view Merging.} Success rate as a function of environment steps averaged over 5 Meta-World hard visual RL tasks. Methods were trained and evaluated on \emph{(Left)} two camera views - First and Third A - \emph{(Right)} three camera views. Dashed line indicates highest performance of singular view baselines. Mean and 95\% CI over 5 random seeds.}
        \label{fig:main_merge}

    \end{minipage}
    \hfill
    \begin{minipage}[t]{0.48\textwidth}
        \centering
        \includegraphics[width=0.92\textwidth,trim={0cm 0.6cm 0 1.5cm},clip]{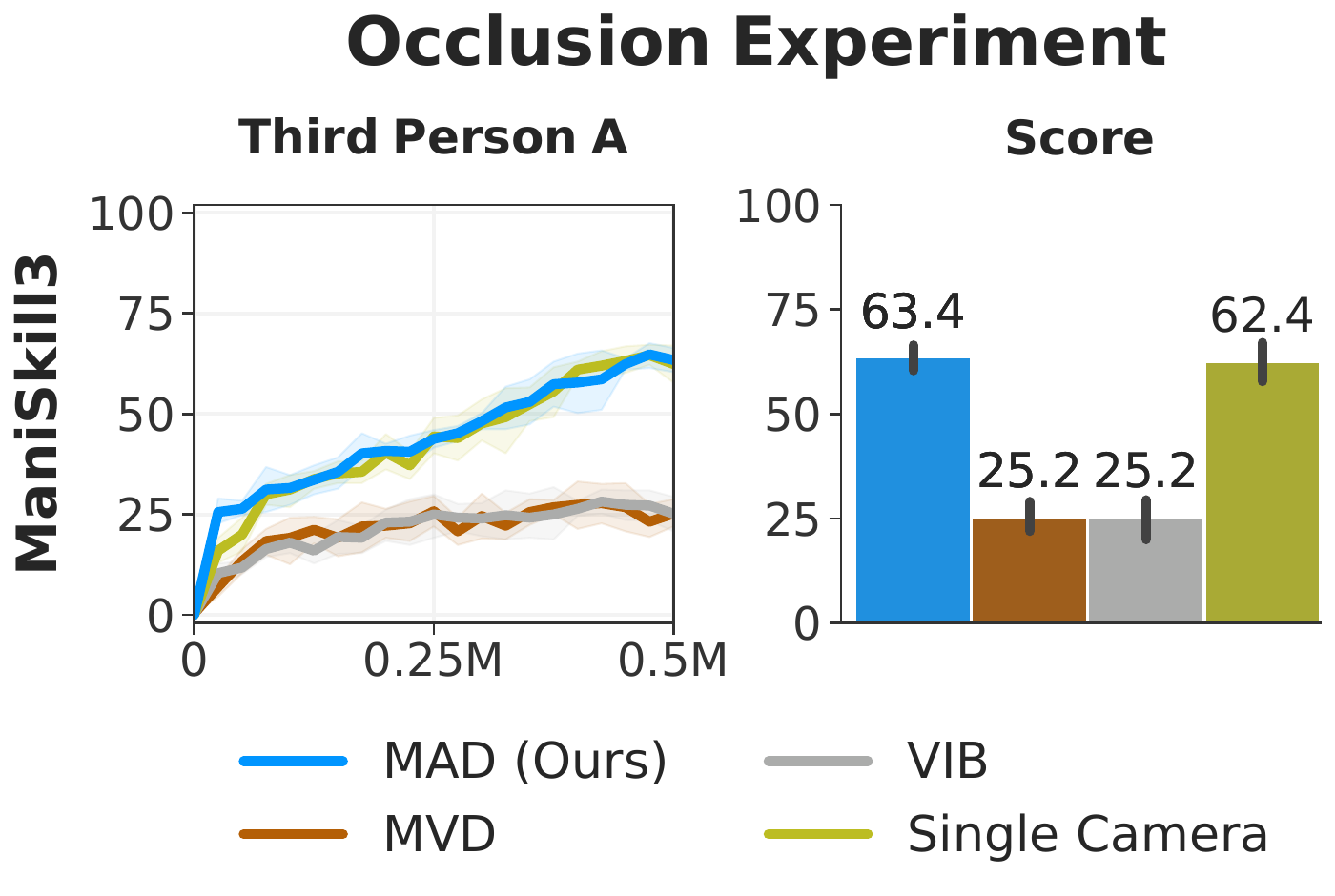}
        \vspace{-0in}
        \caption{\textbf{Occlusion Adaptability.} Success rate as a function of environment steps, averaged over 5 ManiSkill3 tasks. Methods are trained on all three views, with two views fully occluded and only Third Person A being observable. Mean and 95\% CI over 5 random seeds.}
        \label{fig:main_occ}
    \end{minipage}
    \vspace{-0.2in}
\end{figure}

Furthermore, we \emph{compare different merge methods} with our merge module. We train DrQ agents on 5 hard tasks from Meta-World using different merge methods for multi view inputs. These merge techniques consist of early, middle, and late merging. For early merging we employ: 1) Frame Stack, where multiple views are stacked across the channel dimension before being passed to the CNN encoder \cite{mnih2013playing, mnih2015human}. For middle merging we include: 2) Concat, where singular views are passed through the encoder and their features are then concatenated \cite{hsu2022vision} 3) Sum (Ours), where singular view features are summed instead of being concatenated \cite{lancaster2024modem}, 
4) Attention, where cross attention is used between singular view features \cite{jangir2022look}, 5) ViT, where singular view features are passed through a ViT layer to leverage attention across views \cite{seo2023multi}. For late merging we add: 6) Q-mean, where we predict a separate Q-value for each view and then average across all views to get the final Q-value \cite{akinola2020learning, jang2023murm}. We train these merge strategies using our DrQ baseline on two views and on three views and display the results in Figure \ref{fig:main_merge}. The results indicate that all merge methods achieve similar sample efficiency on two and three views. The only merging method that slightly under performs is Q-Mean, which indicates that late merging might not be as effective as other merging strategies for multi view RL.

\textbf{Adaptability.}
To test the \emph{adaptability of MAD to occluded or partial views}, we alter the angles of two cameras to face uninformative views and align only one camera, namely Third Person A, to face the robot arm and task. We train all methods using this setup on 5 ManiSkill3 tasks and report the results in Figure \ref{fig:main_occ}. Our method, MAD, is able to solve the tasks with similar success rate (63\%) as the Single Camera (62\%), despite it being overloaded with uninformative views. Compared to MAD, neither the MVD nor the VIB baselines seem to achieve similar sample efficiency. For the MVD baseline, the three views don't have a useful shared representation, since only one view is observant of the task, and thus it fails to solve the tasks effectively. On the other hand, the VIB baseline fails to solve the tasks due to its heavy dependency on the first person camera view.

\begin{figure*}[t]
    \centering
    \begin{minipage}{0.19\textwidth}
        \centering
        \includegraphics[width=0.8\textwidth,trim={0cm 0 0 0},clip]{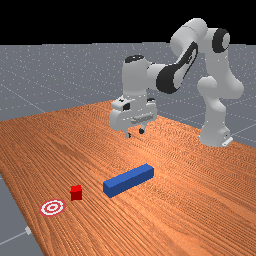}
    \end{minipage}
    \begin{minipage}{0.19\textwidth}
        \centering
        \includegraphics[width=0.8\textwidth,trim={0cm 0 0 0},clip]{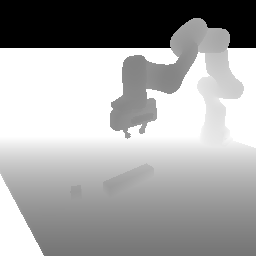}
    \end{minipage}
    \hfill
    \begin{minipage}{0.59\textwidth}    
        \includegraphics[width=1.0\textwidth,trim={0.1cm 0 0 1.5cm},clip]{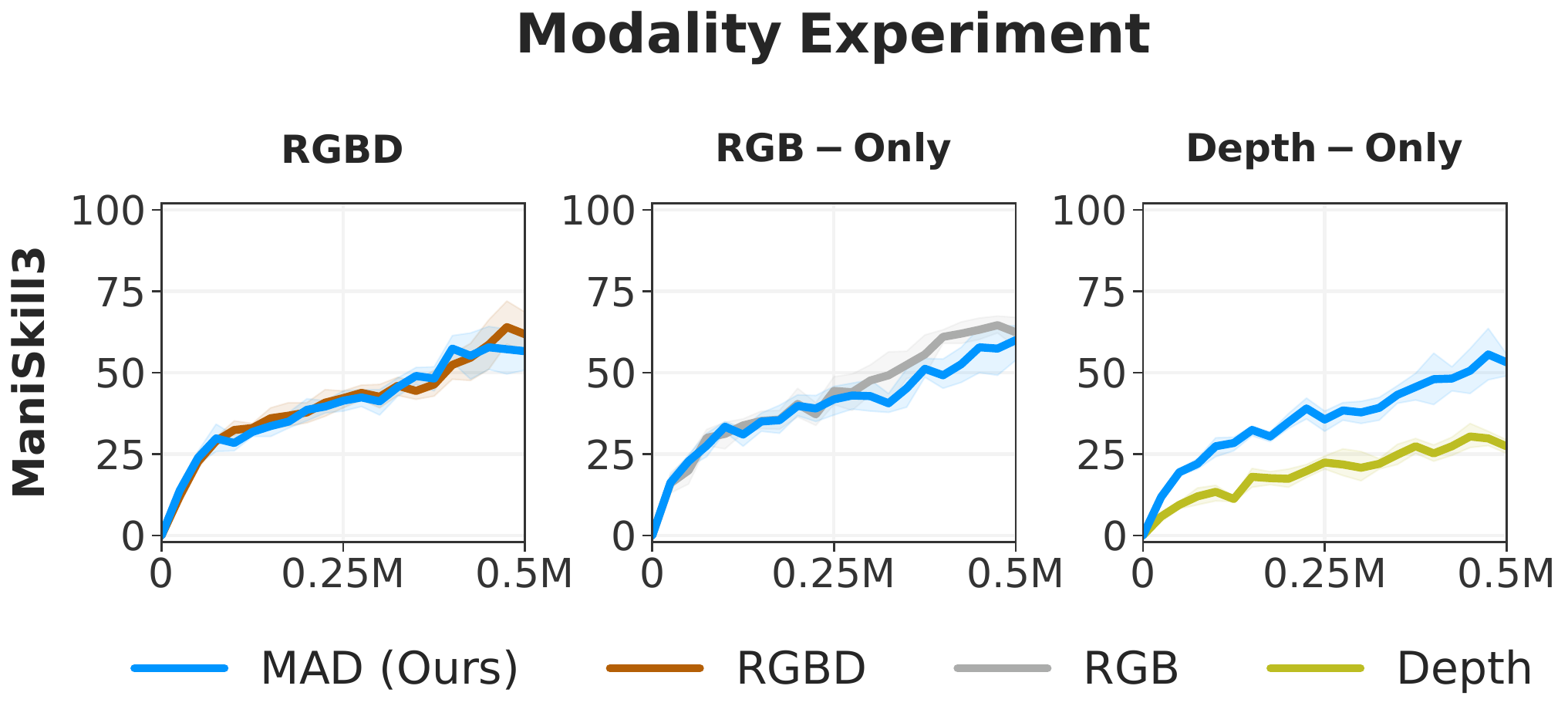}
    \end{minipage}
    \vspace{-0.05in}
    \caption{\textbf{Modality Adaptability.} Success rate as a function of environment steps averaged over 5 ManiSkill3 tasks. \emph{(Left)} RGB and Depth images from the PokeCube Third Person A view. \emph{(Right)} Methods are trained on one camera view (Third Person A) but with different modalities, and evaluated accordingly. Mean and 95\% CI over 5 random seeds. }
    \label{fig:main_mod}
    \vspace{-0.1in}
\end{figure*}

We further evaluate whether \emph{MAD can be adapted to different modalities} such as depth and RGB as opposed to different camera views, and display the results in Figure \ref{fig:main_mod}. We train MAD and DrQ agents on 5 ManiSkill3 tasks, with only one camera view input. We pass in RGBD inputs to MAD, and different combinations of RGB and Depth to DrQ agents to measure the baseline performance on that modality. While RGBD input is usually merged by adding depth as an extra channel to the CNN input, we merge it in MAD by summing its features after encoding them separately, for consistency with the MAD framework. Based on the results, MAD is able to achieve similar sample efficiency to the RGBD baseline, while generalizing to both the RGB-Only and Depth-Only modalities. Most surprisingly, MAD is more sample efficient than the Depth baseline on the Depth-Only evaluation, mainly due to it leveraging more informative representations from the RGB inputs. This can prove to be an effective way to train deployable depth-only policies from RGBD inputs.

\section{Limitations and Conclusion}

A limitation of the current study is the absence of real-world experimental validation.  Future research could address this gap by implementing MAD on physical robots through simulation-to-real transfer methods \cite{lin2025simtoreal}, demonstration-bootstrapped methods \cite{hu2023imitation}, or enhanced sample efficient visual Q-learning methods \cite{xu2023drm, hansen2024tdmpc2, fujimoto2025mrq}. 
Furthermore, generalizing to novel unseen viewpoints in both MAD, and visual RL in general, remains challenging. Future research is needed to expand on the potential of MAD for tackling novel view points, especially with the aid of generative foundational models \cite{van2024generative}, to deploy robust robot policies to the real world.

To conclude, throughout this work, we identified that the two directions in visual reinforcement learning of: \emph{(1)} merging views and \emph{(2)} disentangling views, can be complementary. We proposed our framework, Merge And Disentangle (\textbf{MAD}), which bridges these two directions, benefiting from the merits of each. Concretely in MAD, multi-view inputs are merged for higher sample efficiency from multi-view representations, while singular view features are selectively applied as feature-level augmentations to the downstream components to ensure disentanglement. This produces policies that are robust to a reduction of available cameras, whether in training or deployment. We heavily benchmarked MAD on 20 visual RL tasks from Meta-World and ManiSkill3 against multiple baselines, showcasing its superiority in terms of sample efficiency and robustness. Furthermore, we showed that the application of MAD could be versatile, whether to multiple camera views, or multiple input modalities. Bringing this work forward, we hope this can aid in democratizing visual reinforcement learning, making it more accessible, and serve as a strong baseline for future work in multi view reinforcement learning.


\clearpage

\acknowledgments{This research has in part been sponsored by Kuwait University through a visiting fellow program and by ARL DCIST. The support is gratefully acknowledged.}


\bibliography{main.bib}

\appendix
\onecolumn

\section{Experiment Setup}

\subsection{Hyper-parameters}
\label{subsection:app_hparams}

\begin{table}[H]
\centering
\begin{tabular}{lc}
\toprule
Parameter        & Setting \\
\midrule
Replay buffer capacity & 500,000 \\
Image Size & (3, 84, 84) \\
Frame stack & $3$ \\
Exploration steps & $2000$ \\
Mini-batch size & $256$ \\
Optimizer & Adam \\
Learning rate & $5\times10^{-4}$ \\
Agent update frequency & $2$ \\
Critic Q-function soft-update rate $\tau$ & $0.01$ \\
Features dim. & $50$ \\
Hidden dim. & $1024$ \\
Actor log stddev bounds & $[-10,2]$ \\
Init temperature & $0.1$ \\
MAD Alpha $\alpha$ & $0.8$ \\
Discount $\gamma$ & (Meta-World) $0.99$\\ & (ManiSkill3) $0.8$ \\
\bottomrule
\end{tabular}
\caption{\label{table:hparams} The default set of hyper-parameters used in our experiments.}
\end{table}

\subsection{Baseline Implementations}
\label{subsection:baseline_imp}

For all baselines, we reimplement them on top of our DrQ baseline for a fair comparison. The MV-MWM baseline however was kept using its model-based baseline, as DrQ is model-free. Nevertheless, MV-MWM was tuned accordingly on both benchmarks.

\subsection{Environment Setups}
\label{subsection:env_setups}

\textbf{Meta-World.} We evaluate on 15 tasks where 5 tasks are chosen from each of medium, hard, and very hard difficulties, defined in \citet{seo2023masked}. Every 25,000 environment steps, we evaluate for 20 episodes and report the mean success rate. 

\begin{figure}[htbp]
    \centering
    \begin{minipage}[b]{0.6\linewidth}
        \centering
        \caption*{\textbf{MetaWorld-v2 Tasks}}
        \vspace{-0.05in}
        \resizebox{\linewidth}{!}{%
        \begin{tabular}{c  c c c c}
           \toprule
           \textbf{Task} & \textbf{Difficulty} & \textbf{Action Dim} & \textbf{Proprioceptive State Dim}\\
           \toprule
           Basketball & Medium & 4 & 4\\
           \midrule
           Hammer & Medium & 4 & 4\\
           \midrule
           Peg Insert Side & Medium & 4 & 4\\
           \midrule
           Soccer & Medium & 4 & 4\\
           \midrule
           Sweep Into & Medium & 4 & 4\\
           \midrule
           Assembly & Hard & 4 & 4\\
           \midrule
           Hand Insert & Hard & 4 & 4\\
           \midrule
           Pick Out Of Hole & Hard & 4 & 4\\
           \midrule
           Pick Place & Hard & 4 & 4\\
           \midrule
           Push & Hard & 4 & 4\\
           \midrule
           Shelf Place & Very Hard & 4 & 4\\
           \midrule
           Disassemble & Very Hard & 4 & 4\\
           \midrule
           Stick Pull & Very Hard & 4 & 4\\
           \midrule
           Stick Push & Very Hard & 4 & 4\\
           \midrule
           Pick Place Wall & Very Hard & 4 & 4\\
           \bottomrule
        \end{tabular}
        }
    \end{minipage}
\end{figure}
\vspace{0.4in}

\textbf{ManiSkill3.} We evaluate on 5 visual tasks from ManiSkill3. Every 20,000 environment steps, we evaluate for 20 episodes and report the mean success rate. We alter the default camera setups of the tasks to accommodate our setup. For both PushCube and PokeCube tasks, we alter the state to include the goal position of the cube to keep consistent with other tasks in the ManiSkill3 environments. 

\begin{figure}[H]
    \centering
    \begin{minipage}[b]{0.6\linewidth}
        \centering
        \caption*{\textbf{ManiSkill3 Tasks}}
        \vspace{-0.1in}
        \resizebox{\linewidth}{!}{%
        \begin{tabular}{c c c c c}
            \toprule
            \textbf{Tasks} & \textbf{Difficulty} & \textbf{Action Dim} & \textbf{Proprioceptive State Dim}\\
            \toprule
            Push Cube & - & 4 & 28 \\
            \midrule
            Pull Cube & - & 4 & 28 \\
            \midrule
            Pick Cube & - & 4 & 29 \\
            \midrule
            Place Sphere & - & 4 & 29\\
            \midrule
            Poke Cube & - & 4 & 28 \\
            \bottomrule
        \end{tabular}
        }
    \end{minipage}
\end{figure}

\begin{figure}[htbp]
    \centering
    \caption*{\textbf{Environment Settings}}
    \vspace{-0.1in}
    \begin{minipage}[b]{\linewidth}
        \centering
        \resizebox{\linewidth}{!}{%
        \begin{tabular}{c c c c c c c}
            \toprule
                & \textbf{Environment Steps} & \textbf{Episode Length} & \textbf{Action Repeat} & \textbf{Metric} & \textbf{Number of Eval Episodes} \\
            \toprule
            \textbf{Meta-World} & 1M & 200 & 2 & Success & 20 \\
            \midrule
            \textbf{ManiSkill3} & 0.5M & 50 & 1 & Success & 20\\
            \bottomrule
        \end{tabular}
        }
    \end{minipage}
\end{figure}

\clearpage
\section{Camera Scalability Experiments}
\label{sec:app_scalability}

\subsection{Increased Camera Views}
We further evaluate how our method would scale with larger number of views. We setup ManiSkill3 with five views as outlined below and evaluate our method and baselines on 5 ManiSkill3 tasks. As shown below, MAD outperforms all baselines by (48\%) success rate when increased to 5 views.  
\begin{figure}[H]
    \centering
    \includegraphics[width=\textwidth,trim={0 0 0 0cm},clip]{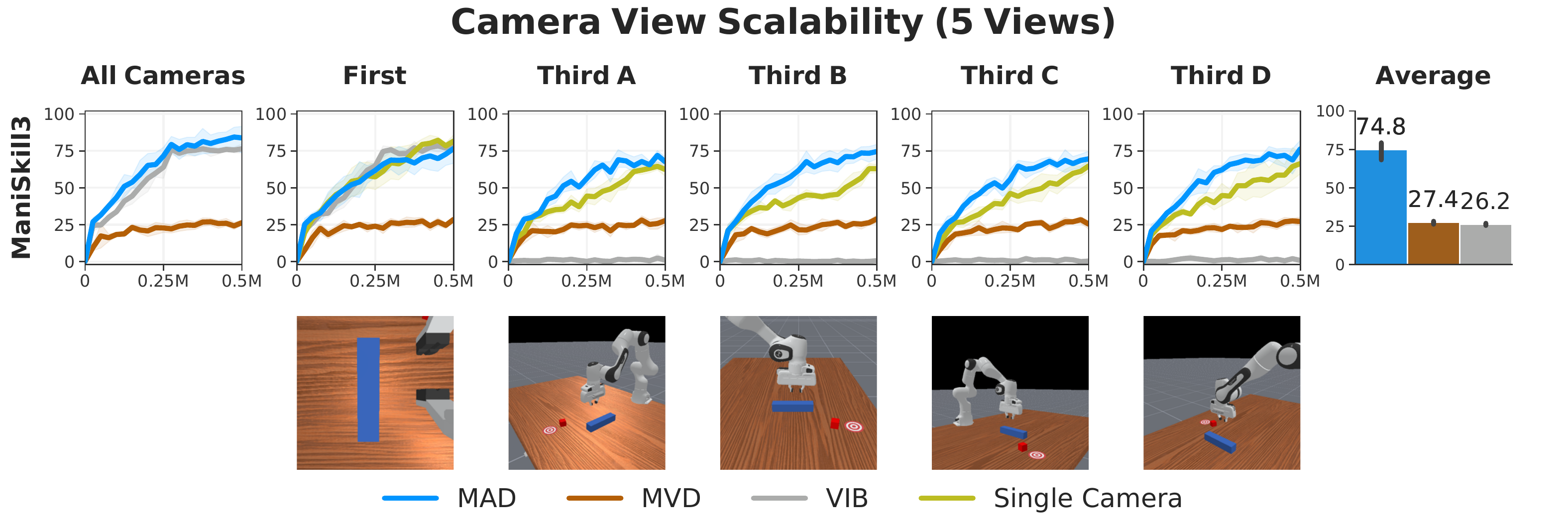}
    \caption{\textbf{View Robustness.} Success rate as a function of environment steps, averaged over 5 ManiSkill3 visual RL tasks. Methods are trained on all five camera views and evaluated on all and singular camera views with the final average displayed on the far right. Mean and 95\% CI over 5 random seeds. }
    \label{fig:five_views}
\end{figure}

\subsection{Different Input Camera View Sizes}

We also evaluate the case where we have multiple camera inputs with different resolutions. If we have multiple cameras with different resolutions, some options to process them are:

(a) As a preprocessing step, resize all images to a fixed size.

(b) Use separate CNN encoders for each image size. This would mean that the input views need to be ordered, such that it is possible to assign each input view to its corresponding CNN encoder.

We opt for option (a) due to its simplicity. Given a First Person View of size (64x64), a Third Person A View of size (96x96) and a Third Person B View of size (84x84), we resize all input images to (84x84) and train MAD on 5 ManiSkill3 tasks to empirically validate our suggestions. We display the results in the figure below, where we find that MAD is able to support different sizes of camera views with a similar performance to our original setup.

\begin{figure}[H]
    \centering
    \includegraphics[width=0.7\textwidth,trim={0 0 0 0cm},clip]{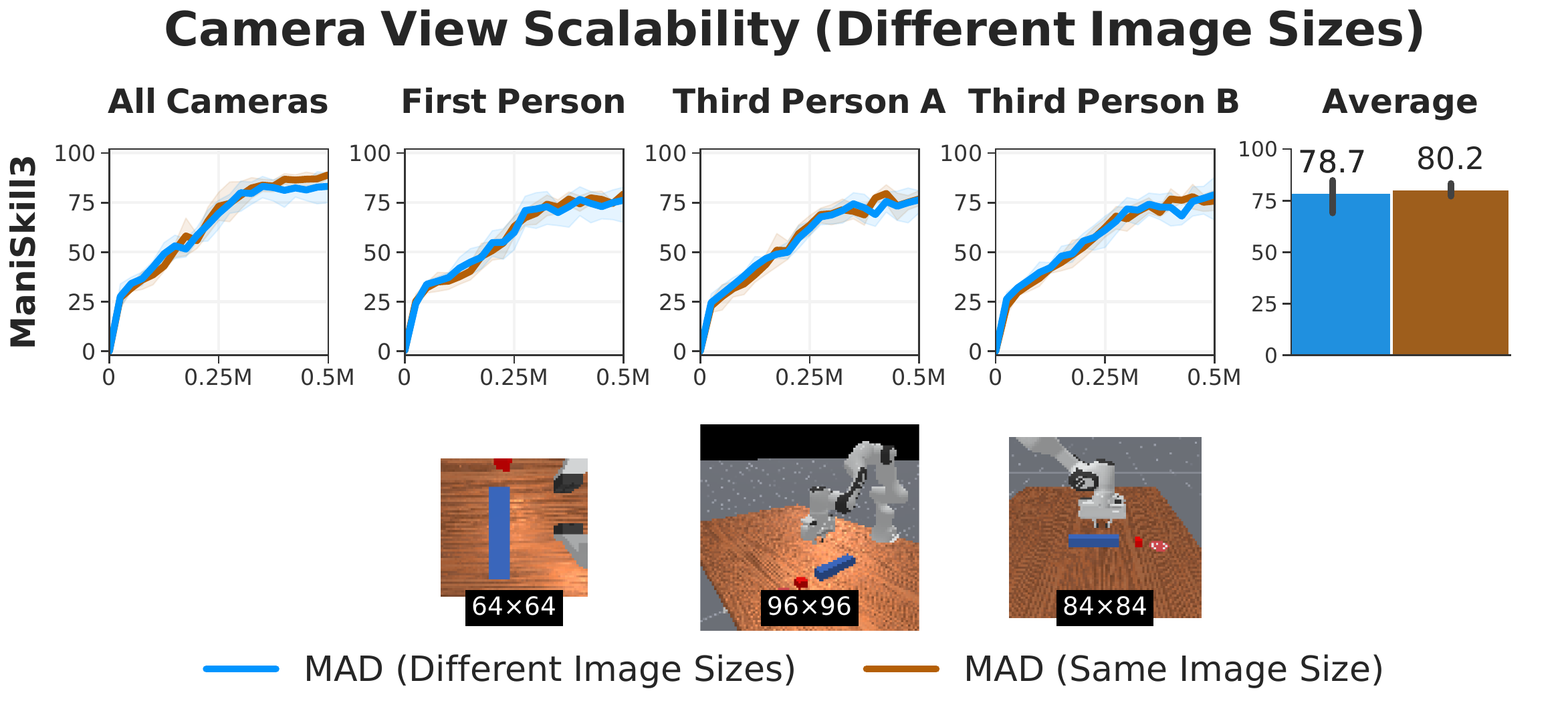}
    \caption{\textbf{View Robustness.} Success rate as a function of environment steps, averaged over 5 ManiSkill3 visual RL tasks. Methods are trained on all three camera views and evaluated on all and singular camera views with the final average displayed on the far right. Mean and 95\% CI over 5 random seeds. }
    \label{fig:diff_image_sizes}
\end{figure}

\clearpage
\section{Extended Analysis}
\label{sec:app_analysis}

\subsection{Best Camera View for Learning}

\begin{figure}[H]
    \centering
    \includegraphics[width=0.6\textwidth,trim={0cm 0 0 2.1cm},clip]{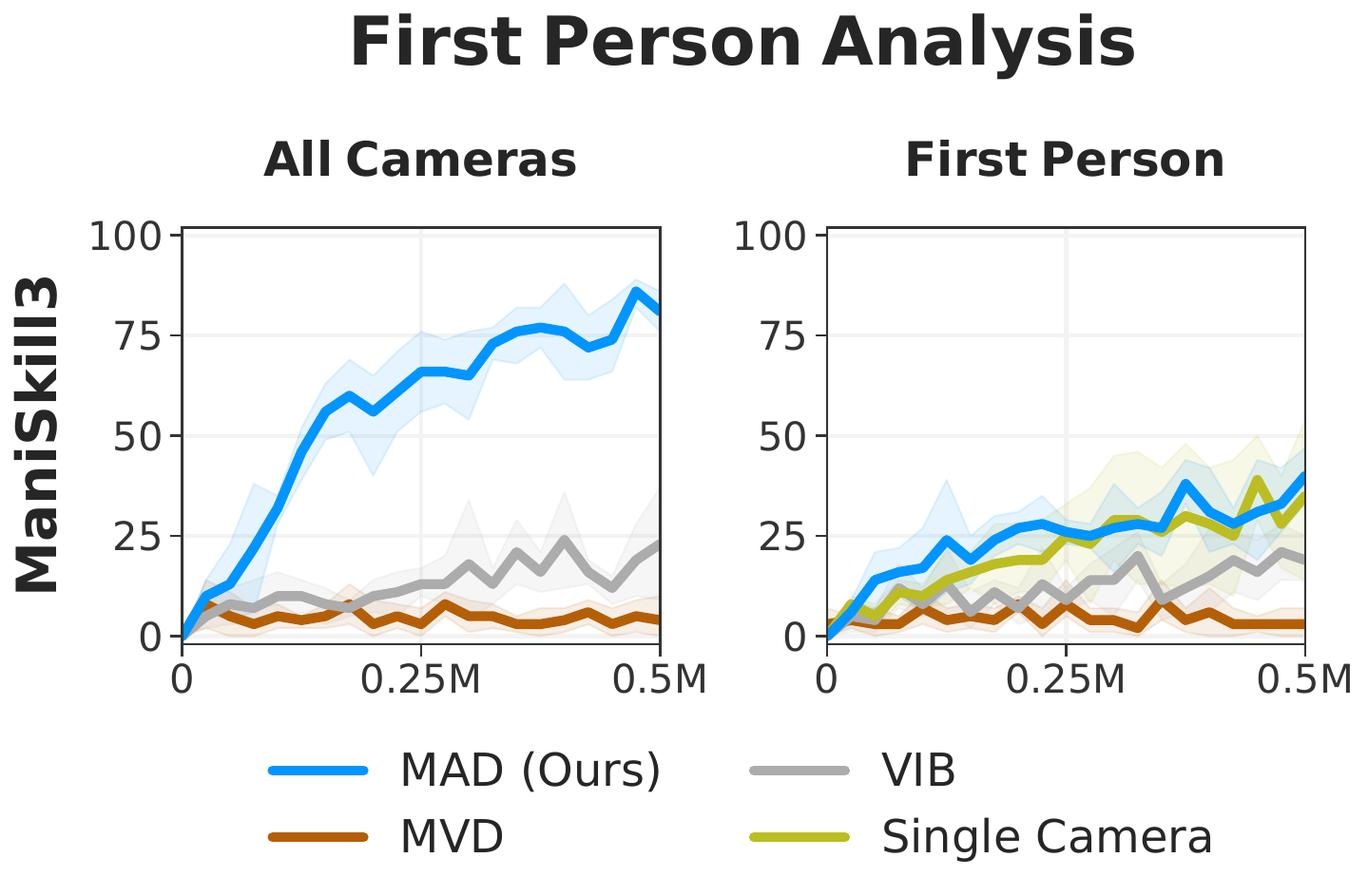}
    \caption{\textbf{First Person Camera View Failure.} Success rate as a function of environment steps on the ManiSkill3 PokeCube task. Methods are trained on all three views and evaluated an all and singular views separately. Mean and 95\% CI over 5 random seeds.}
    \label{fig:main_pokecube_analysis}
\end{figure}

\textbf{Is there a single camera view that consistently outperforms other camera views?} Based on our experiments, \emph{there is a singular view that outperforms other singular views}, and that being the First Person view. This observation aligns with findings from \citet{hsu2022vision}. Due to the first person camera's attachment to the robot arm, and active perception system, it provides the best representations for robot learning. However, its attachment to the robot arm can be a double edged sword, as it can suffer from limited observability that prevents it from solving the task. We shed light on the PokeCube task from ManiSkill3 in Figure \ref{fig:main_pokecube_analysis}, where the First Person view isn't sufficient to solve the task as it has limited observability. Nevertheless, MAD is able to leverage useful information from other third person views and achieve high success rate on All Cameras. In the case of a reduction in views, MAD can still maintain the highest possible success rate on the First Person view given its limited observability.

\clearpage
\section{Extended Results}
\label{sec:app_res}

\subsection{Overall Robustness}
\begin{figure}[H]
    \centering
    \begin{tabular}{c | c}
        \includegraphics[width=0.43\linewidth]{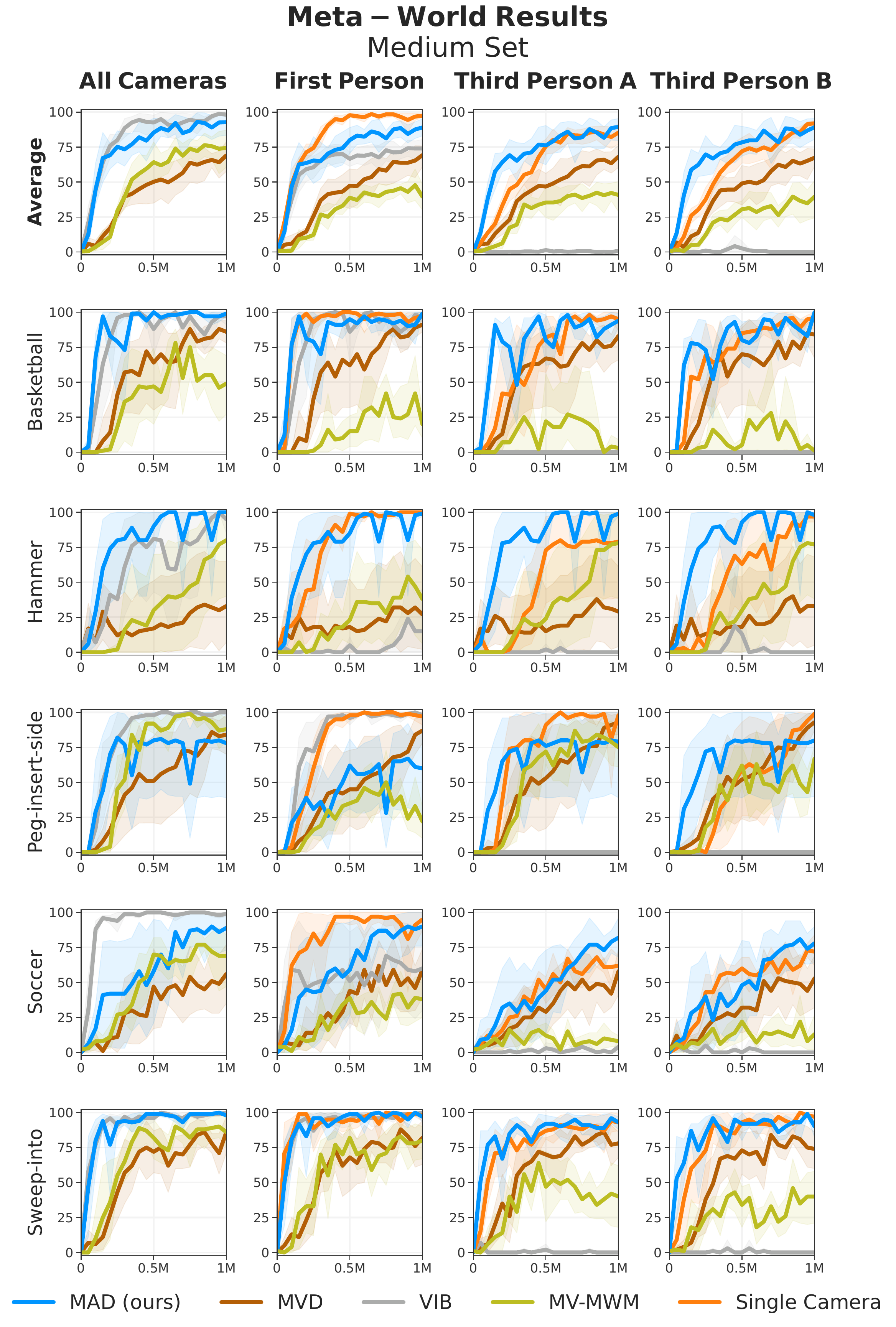} &
        \includegraphics[width=0.43\linewidth]{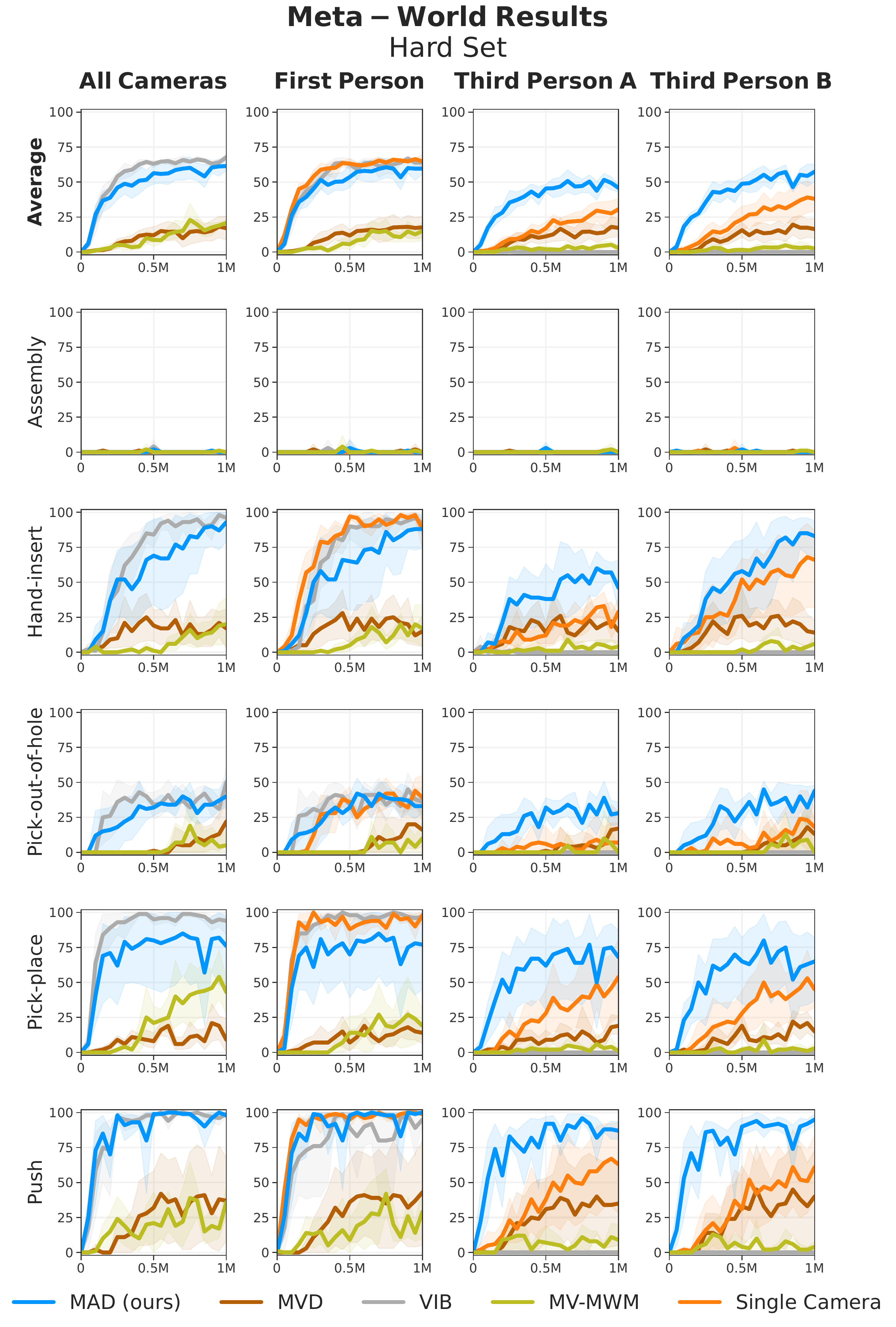} \\ [0.2cm]
        \hline \\
        \includegraphics[width=0.43\linewidth]{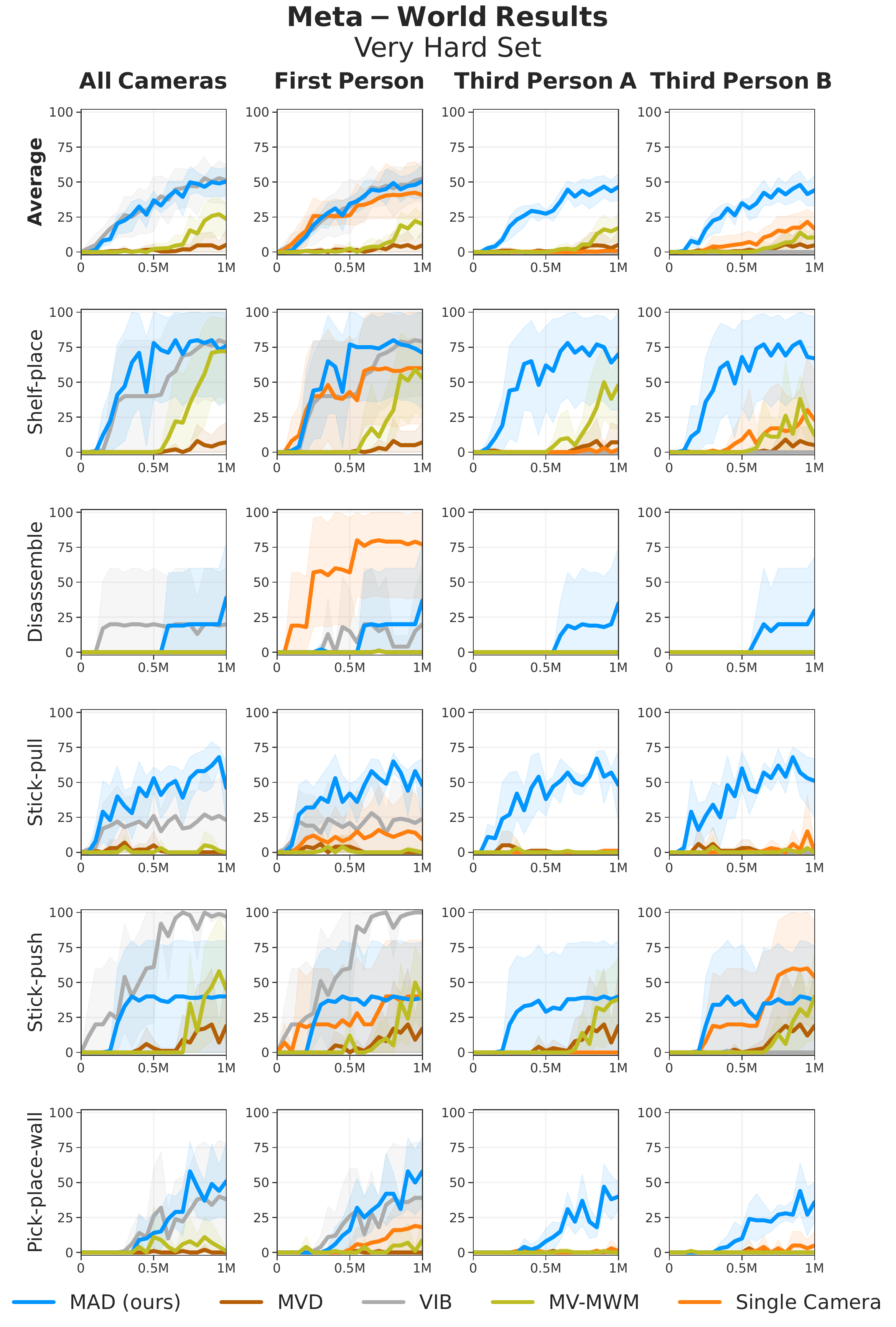} &
        \includegraphics[width=0.43\linewidth]{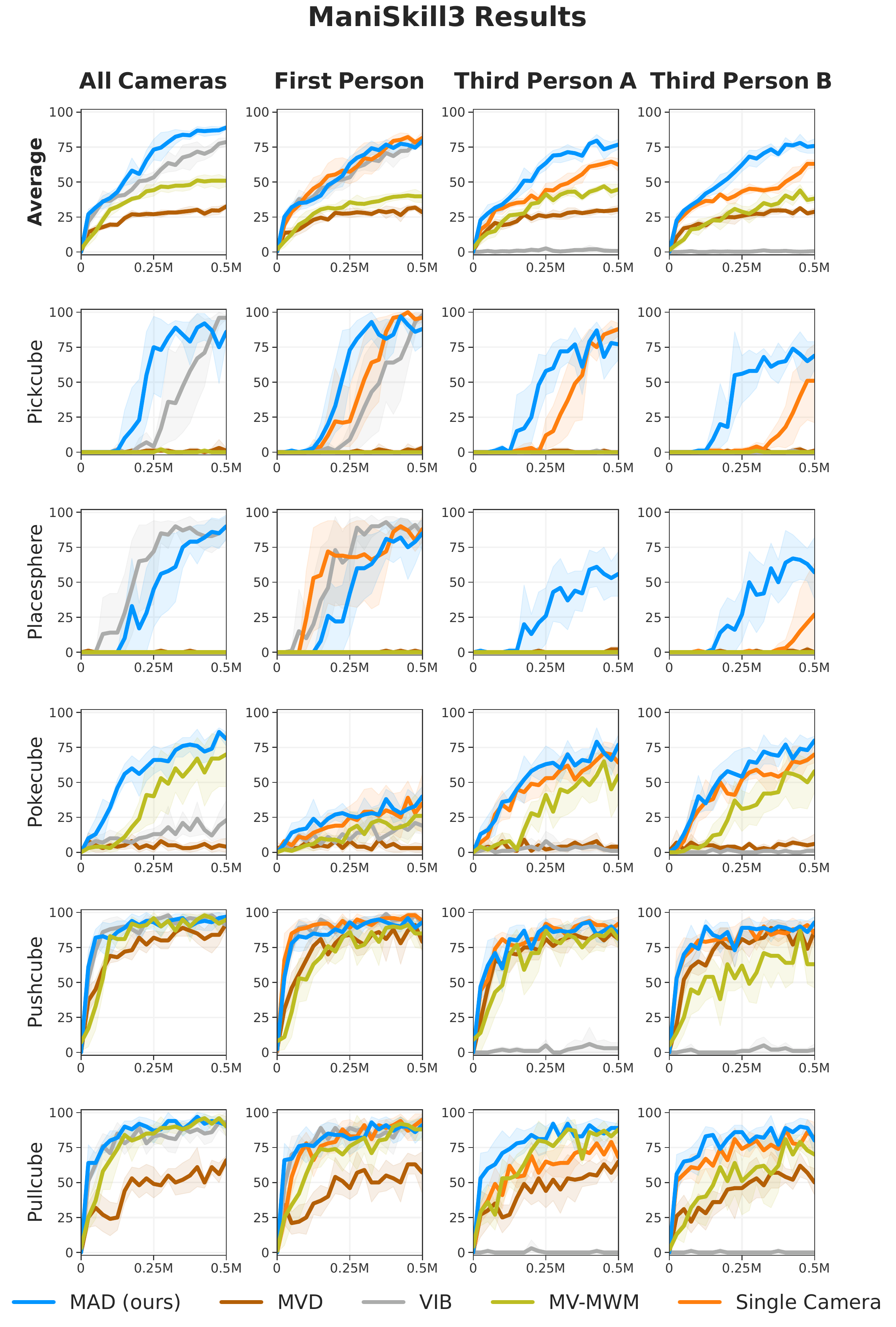}
    \end{tabular}
    \caption{\textbf{Overall Robustness.} Success rate as a function of environment steps, averaged over 15 Meta-World and 5 ManiSkill3 visual RL tasks. \emph{(Top Left)} Meta-World Medium. \emph{(Top Right)} Meta-World Hard. \emph{(Bottom Left)} Meta-World Very Hard. \emph{(Bottom Right)} ManiSkill3. Methods are trained on all three camera views and evaluated on all and singular camera views. Mean and 95\% CI over 5 random seeds. In reference to Figure \ref{fig:main_exp} in the main text.}
\end{figure}

\subsection{Ablations}
\begin{figure}[htbp]
    \centering
    \begin{tabular}{c | c}
        \includegraphics[width=0.47\linewidth]{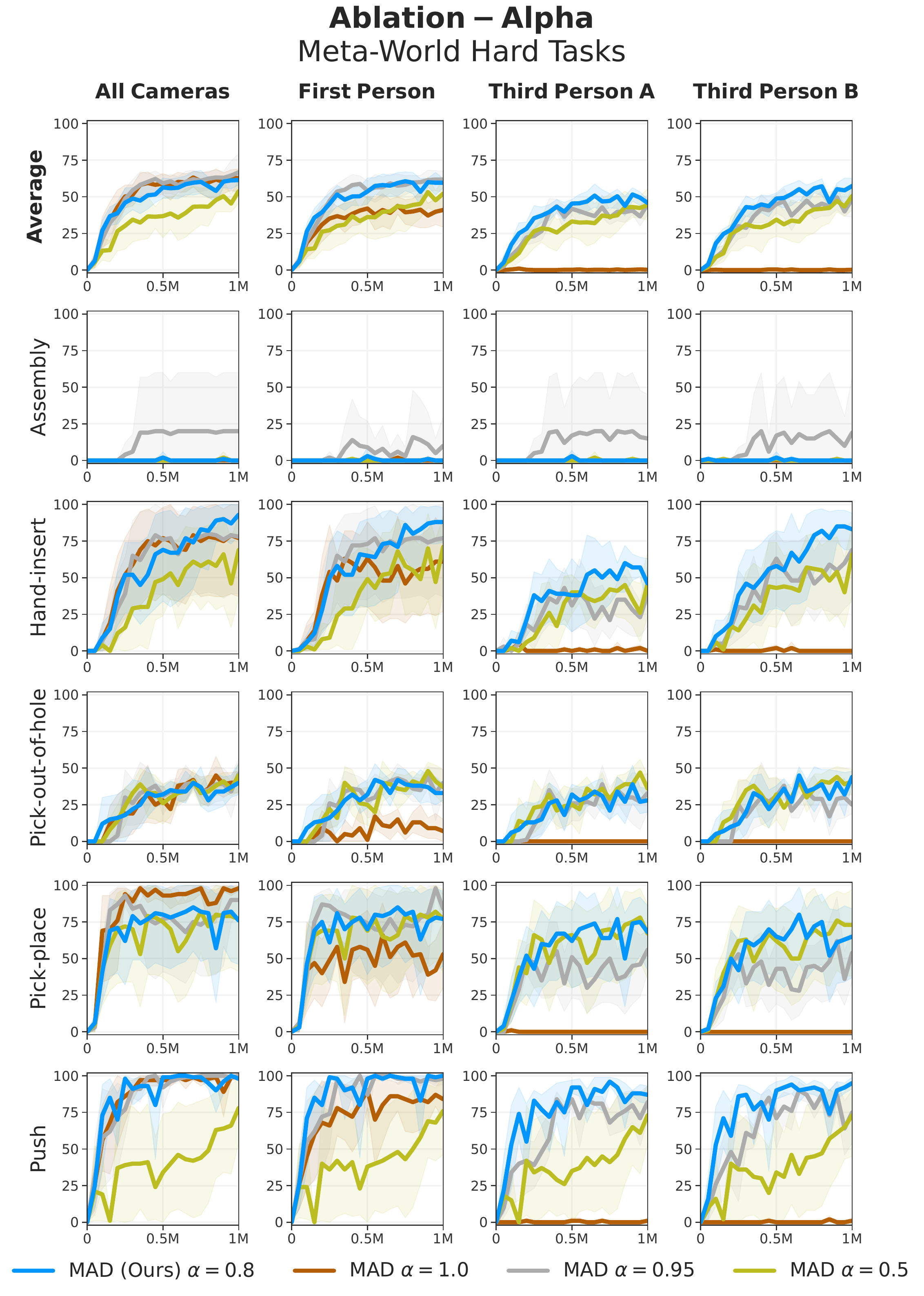} &
        \includegraphics[width=0.53\linewidth]{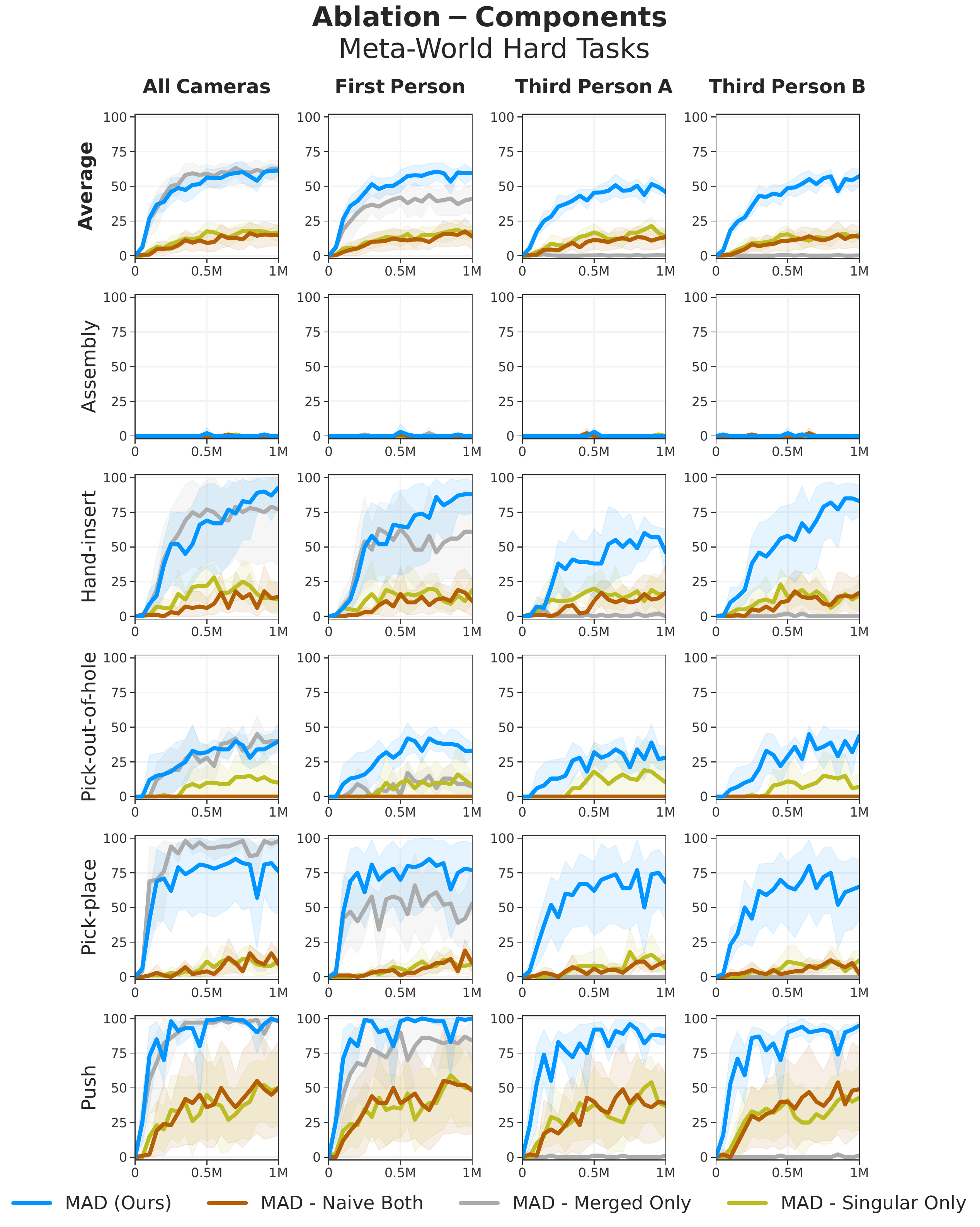} 
    \end{tabular}
    \caption{\textbf{Ablations.} Success rate as a function of environment steps, averaged  over 5 Meta-World hard visual RL tasks. \emph{(Left)} Alpha Ablations. \emph{(Right)} Component Ablations. Methods are trained on all three camera views and evaluated on all and singular camera views. Mean and 95\% CI over 5 random seeds. In reference to Figure \ref{fig:main_abl} in the main text.}
\end{figure}
\clearpage

\subsection{Multi-View Merging}
\begin{figure}[H]
    \centering
    \includegraphics[width=0.8\linewidth]{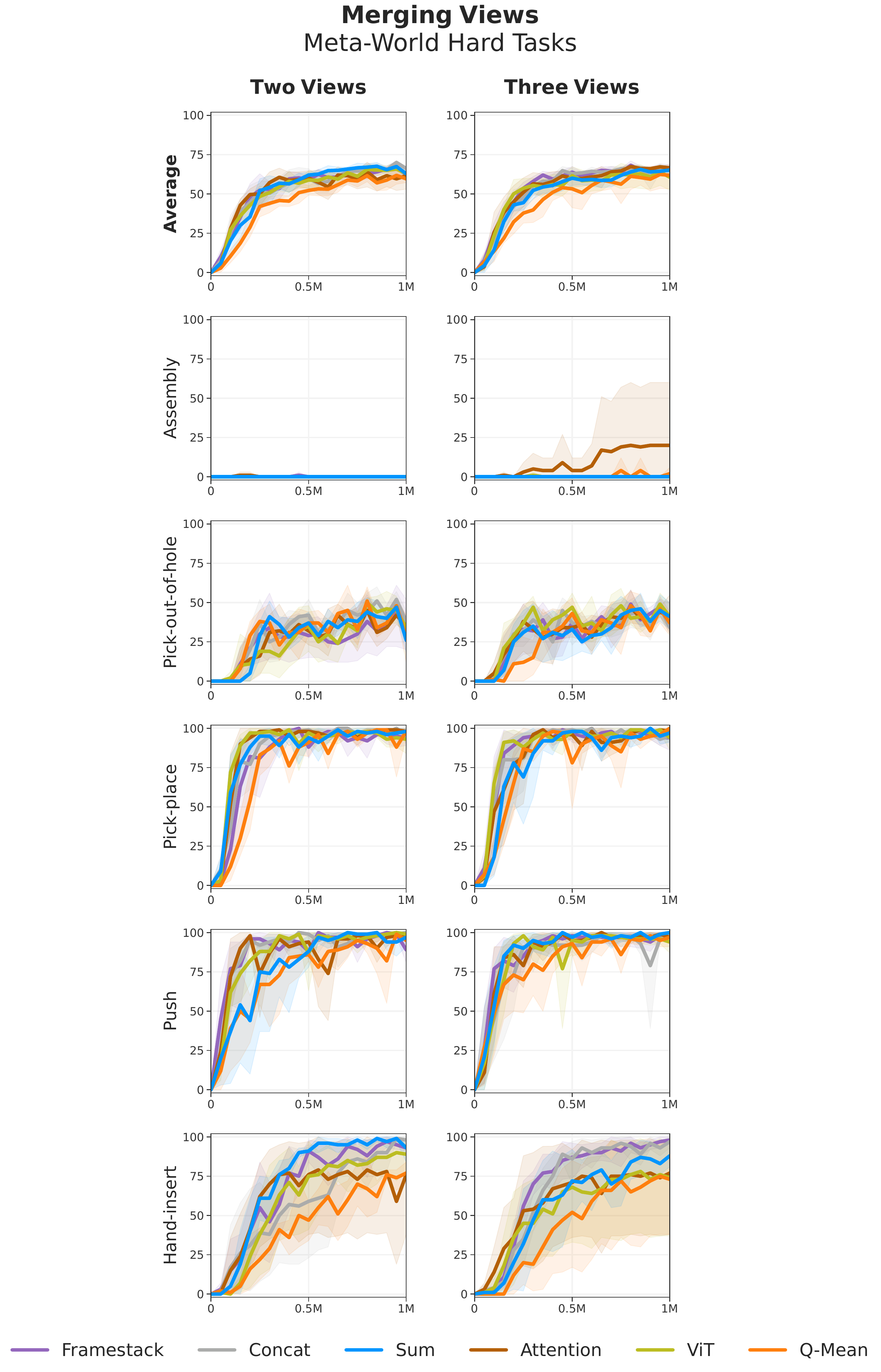}
    \caption{\textbf{Multi-view Merging.} Success rate as a function of environment steps, averaged over 5 Meta-World hard visual RL tasks. Methods trained and evaluated on \emph{(Left)} two camera views - First and Third A - \emph{(Right)} three camera views. Mean and 95\% CI over 5 random seeds. In reference to Figure \ref{fig:main_merge} in the main text.}
    \label{fig:merge_app}
\end{figure}

\clearpage
\section{Visuals}
\label{sec:app_visuals}

 \input{figs/env_visuals}


\end{document}

%% file: figs/env_visuals.tex
\newcommand{\tasktable}[2]{
\begin{table}[H]
\renewcommand{\thetable}{}
\renewcommand{\tablename}{}
\centering
\textbf{#1} \\
\smallskip
\smallskip
\smallskip
\setlength{\arrayrulewidth}{1pt}
\begin{tabular}{>{\centering\arraybackslash}m{2.0cm}*{5}{>{\centering\arraybackslash}m{2.0cm}}}
\toprule
\textbf{View} & #2 \\
\midrule
First View & \includegraphics[width=0.15\textwidth]{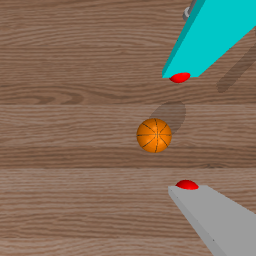} & 
\includegraphics[width=0.15\textwidth]{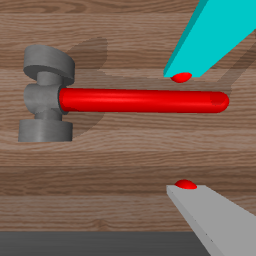} &
\includegraphics[width=0.15\textwidth]{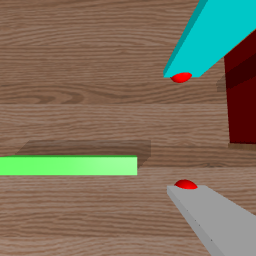} &
\includegraphics[width=0.15\textwidth]{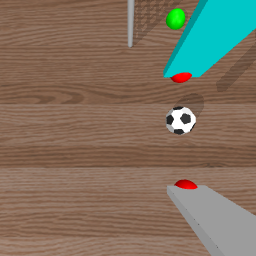} &
\includegraphics[width=0.15\textwidth]{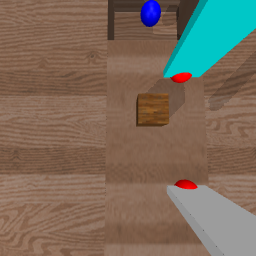} \\[0.2cm]
Third View A & \includegraphics[width=0.15\textwidth]{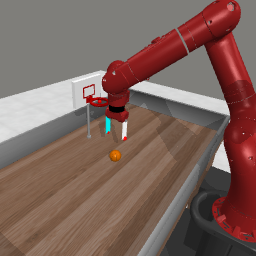} &
\includegraphics[width=0.15\textwidth]{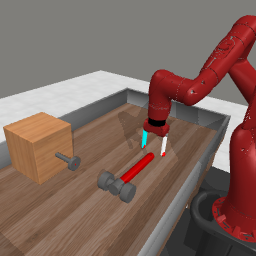} &
\includegraphics[width=0.15\textwidth]{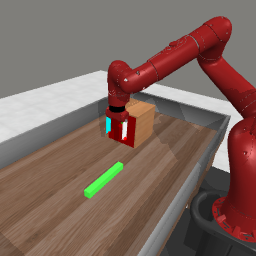} &
\includegraphics[width=0.15\textwidth]{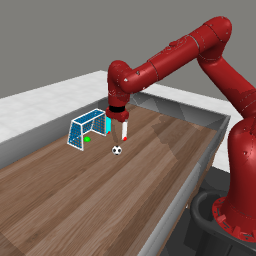} &
\includegraphics[width=0.15\textwidth]{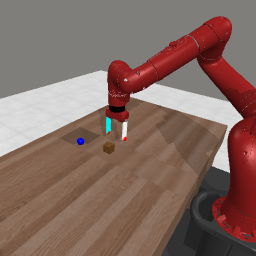} \\[0.2cm]
Third View B & \includegraphics[width=0.15\textwidth]{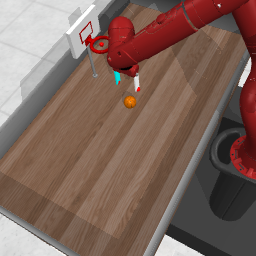} &
\includegraphics[width=0.15\textwidth]{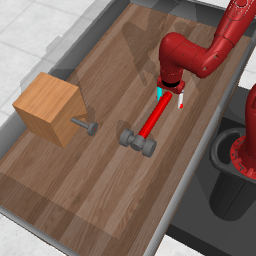} &
\includegraphics[width=0.15\textwidth]{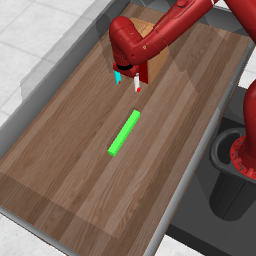} &
\includegraphics[width=0.15\textwidth]{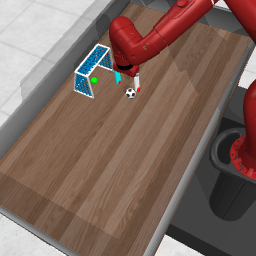} &
\includegraphics[width=0.15\textwidth]{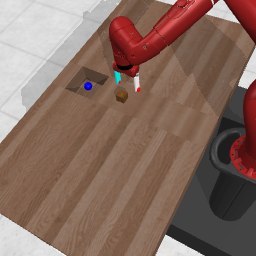} \\
\bottomrule
\end{tabular}
\end{table}
\vspace{-0.4in}
}

\subsection{Meta-World}
\vspace{-0.3in}

\def\tasknameA{basketball}
\def\tasknameB{hammer}
\def\tasknameC{peg-insert-side}
\def\tasknameD{soccer}
\def\tasknameE{sweep-into}
\tasktable{Medium Tasks}{Basketball & Hammer & Peg Insert Side & Soccer & Sweep Into}

\def\tasknameA{assembly}
\def\tasknameB{hand-insert}
\def\tasknameC{pick-out-of-hole}
\def\tasknameD{pick-place}
\def\tasknameE{push}
\tasktable{Hard Tasks}{Assembly & Hand Insert & Pick Out Of Hole & Pick Place & Push}

\def\tasknameA{shelf-place}
\def\tasknameB{disassemble}
\def\tasknameC{stick-pull}
\def\tasknameD{stick-push}
\def\tasknameE{pick-place-wall}
\tasktable{Very Hard Tasks}{Shelf Place & Disassemble & Stick Pull & Stick Push & Pick Place Wall}

\vspace{2cm}
\subsection{ManiSkill3}
\vspace{1cm}

\def\tasknameA{PickCube}
\def\tasknameB{PlaceSphere}
\def\tasknameC{PokeCube}
\def\tasknameD{PushCube}
\def\tasknameE{PullCube}
\tasktable{ManiSkill3 Tasks}{Pick Cube & Place Sphere & Poke Cube & Push Cube & Pull Cube}